%% file: main.tex
\documentclass{article}

\makeatletter
\def\@fnsymbol#1{\ensuremath{\ifcase#1\or \dagger\or \ddagger\or
   \mathsection\or \mathparagraph\or \|\or **\or \dagger\dagger
   \or \ddagger\ddagger \else\@ctrerr\fi}}
\makeatother
\PassOptionsToPackage{numbers, compress}{natbib}

\usepackage[final]{neurips_2024}

\usepackage[utf8]{inputenc} 
\usepackage[T1]{fontenc}    
\usepackage{hyperref}       
\usepackage{url}            
\usepackage{booktabs}       
\usepackage{amsfonts}       
\usepackage{nicefrac}       
\usepackage{microtype}      
\usepackage{xcolor}         

\usepackage{paralist}
\usepackage{xspace}

\def\ourName{SFD\xspace}
\title{Simple and Fast Distillation of Diffusion Models}

\author{%
  Zhenyu Zhou$^{1}$\quad Defang Chen$^{2}$\thanks{Corresponding author. Work partially done during Defang's time at Zhejiang University.} \quad Can Wang$^{1}$\quad Chun Chen$^{1}$\quad Siwei Lyu$^{2}$ \\
  $^1$Zhejiang University
  \quad
  $^2$University at Buffalo, State University of New York \\
  \texttt{\{zhyzhou, defchern\}@zju.edu.cn}
}

\input{cmd.tex}

\begin{document}

\maketitle

\input{sec/abstract}
\input{sec/intro}
\input{sec/preliminary}

\input{sec/method}

\input{sec/exp}

\input{sec/conclusion}

\bibliographystyle{plain}
\bibliography{main}

\input{sec/appendix}

\end{document}

%% file: cmd.tex
\usepackage{amsmath}
\usepackage{amssymb}
\usepackage{amsthm}
\usepackage{booktabs}
\usepackage{subcaption}
\usepackage{mathtools}
\usepackage{pifont}
\usepackage{cancel}
\usepackage{algorithm}
\usepackage{algorithmic}
\usepackage{multirow}
\usepackage{colortbl}
\usepackage{bbding}
\usepackage{utfsym}
\usepackage{xcolor}
\usepackage{tikz}
\usepackage{graphicx}
\usepackage{wrapfig}

\def\bff{\mathbf{f}}

\def\bfw{\mathbf{w}}
\def\bfx{\mathbf{x}}

\def\bfI{\mathbf{I}}

\def\bfeps{\boldsymbol{\epsilon}}

\def\nablax{\nabla_{\bfx}}

\def\rmd{\mathrm{d}}
\def\rmdx{\mathrm{d}\bfx}

\def\bbR{\mathbb{R}}

\def\calN{\mathcal{N}}

\def\eqref#1{Eq.~(\ref{#1})}


%% file: sec/abstract.tex
\begin{abstract}
Diffusion-based generative models have demonstrated their powerful performance across various tasks, but this comes at a cost of the slow sampling speed. To achieve both efficient and high-quality synthesis, various distillation-based accelerated sampling methods have been developed recently. However, they generally require time-consuming fine tuning with elaborate designs to achieve satisfactory performance in a specific number of function evaluation (NFE), making them difficult to employ in practice. To address this issue, we propose \textbf{S}imple and \textbf{F}ast \textbf{D}istillation (SFD) of diffusion models, which simplifies the paradigm used in existing methods and largely shortens their fine-tuning time up to $1000\times$. We begin with a vanilla distillation-based sampling method and boost its performance to state of the art by identifying and addressing several small yet vital factors affecting the synthesis efficiency and quality. Our method can also achieve sampling with variable NFEs using a single distilled model. Extensive experiments demonstrate that SFD strikes a good balance between the sample quality and fine-tuning costs in few-step image generation task. For example, SFD achieves 4.53 FID (NFE=2) on CIFAR-10 with only \textbf{0.64 hours} of fine-tuning on a single NVIDIA A100 GPU. Our code is available at \url{https://github.com/zju-pi/diff-sampler}.
\end{abstract}

%% file: sec/intro.tex
\section{Introduction}
\label{sec:intro}

\begin{wrapfigure}{r}{0.5\textwidth}
  \vspace{-\intextsep}     
  \centering
  \includegraphics[width=0.45\textwidth]{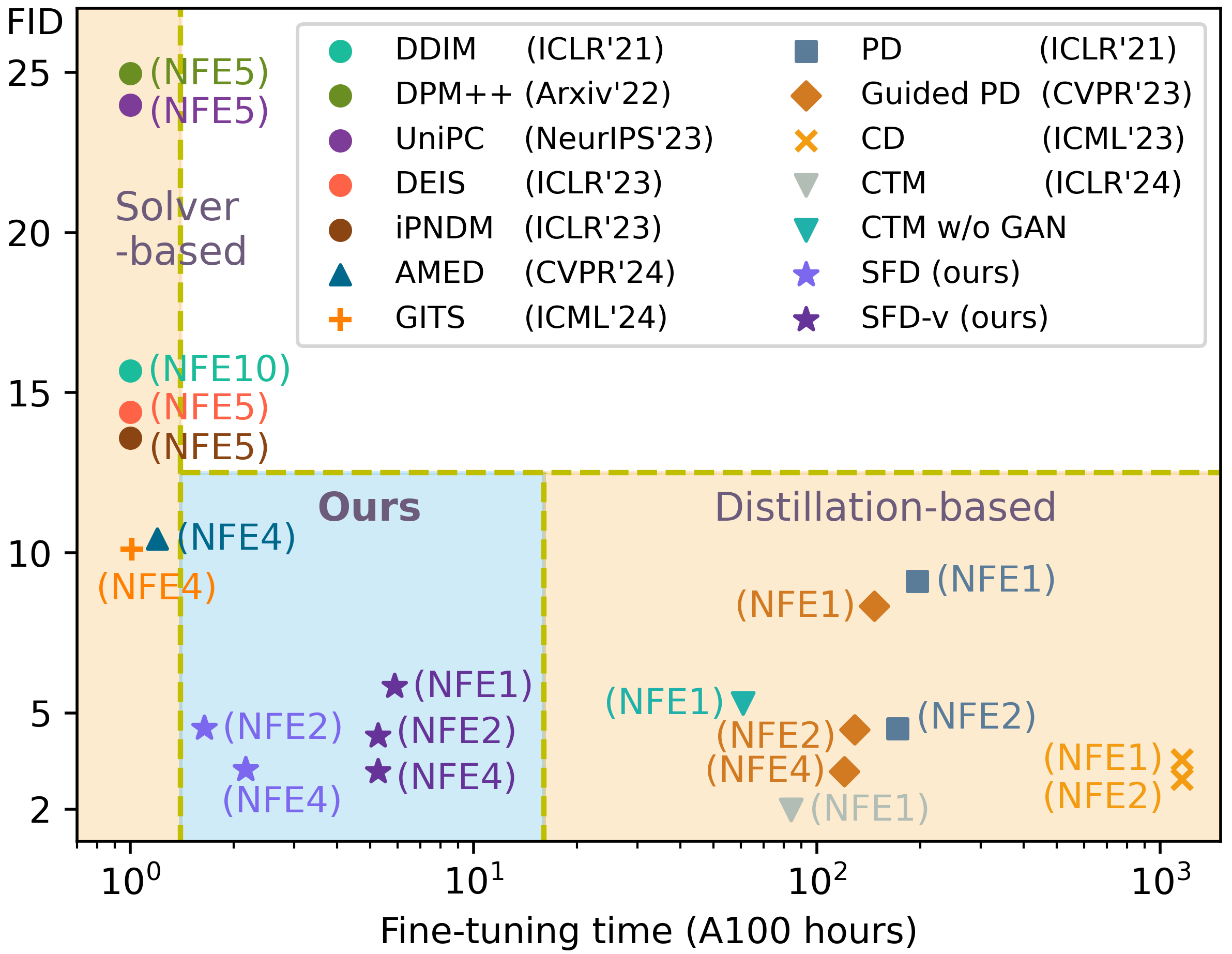}
  \caption{\it \small 
  Comparison of acceleration methods on diffusion models. For better visualization, the time axis is shifted by adding one hour to the actual time required.   
  Our method achieves good performance with a small fine-tuning cost.
  Note that it takes about {\bf 200 hours} to train a diffusion model from scratch in this setting.
  }
  \label{fig:teaser}
  \vspace{-\intextsep}
\end{wrapfigure}

Diffusion models have attracted increasing interest in recent years due to their remarkable generative abilities across various domains, including image~\cite{rombach2022ldm,saharia2022photorealistic,ruiz2023dreambooth}, video~\cite{ho2022video,blattmann2023videoLDM}, audio~\cite{kong2021diffwave,liu2023audioldm}, and molecular structures~\cite{xu2022geodiff}. These models progressively transform a noisy input into a realistic output through iterative denoising steps.
Diffusion models are preferred over other generative models~\cite{goodfellow2014generative,kingma2013auto} for their high-quality synthesis, stable training and a strong theoretical foundation rooted in stochastic differential equations~\cite{song2021sde}. However, achieving high-quality synthesis with diffusion models typically requires hundreds to thousands of sampling steps, resulting in slow sampling speeds and a significant challenge for practical applications.

Recent years have witnessed significant progress in accelerating the sampling of diffusion models~\cite{song2021ddim,lu2022dpmpp,zhang2023deis,zhao2023unipc,gonzalez2024seeds,pandey2023efficient,zhou2023fast,chen2024trajectory,luhman2021knowledge,salimans2022progressive,meng2023distillation,song2023consistency,kim2023consistency,sun2023accelerating,feng2024relational}. These methods typically fall into two categories: \textit{solver-based methods} and \textit{distillation-based methods}. Solver-based methods~\cite{song2021ddim,lu2022dpmpp,zhang2023deis,karras2022edm,zhao2023unipc,gonzalez2024seeds,pandey2023efficient,zhou2023fast,chen2024trajectory} consider sampling from diffusion models as solving differential equations, and employ fast numerical solvers to accelerate high-quality synthesis. However, these methods are limited by inherent truncation errors, and the sample quality becomes degraded when the number of function evaluations (NFEs) is relatively small (e.g., NFE $\leq 5$). 
Distillation-based methods, on the other hand, retain the structure of the original (teacher) diffusion model but aim to create a simplified (student) model that streamlines the iterative refinement process of diffusion models~\cite{luhman2021knowledge,salimans2022progressive,meng2023distillation,song2023consistency,kim2023consistency,sun2023accelerating,feng2024relational}. Extreme distillation-based methods even establish a direct one-to-one mapping between the implicit data distribution and a pre-specified noise distribution~\cite{luhman2021knowledge,liu2022flow,song2023consistency,gu2023boot,yin2023one}. Although distillation-based methods have demonstrated impressive results, often outperforming solver-based methods in sampling quality given the total NFE budge is less than 5, they require expensive computational resources to fine tuning pre-trained diffusion models. As illustrated in Figure \ref{fig:teaser}, the necessary time generally exceeds one hundred GPU hours, \textit{even reaching the same order of magnitude required for training a diffusion model from scratch}. We attribute the time-consuming fine-tuning process to the following two factors:

\begin{figure}
  \centering
  \includegraphics[width=1.0\linewidth]{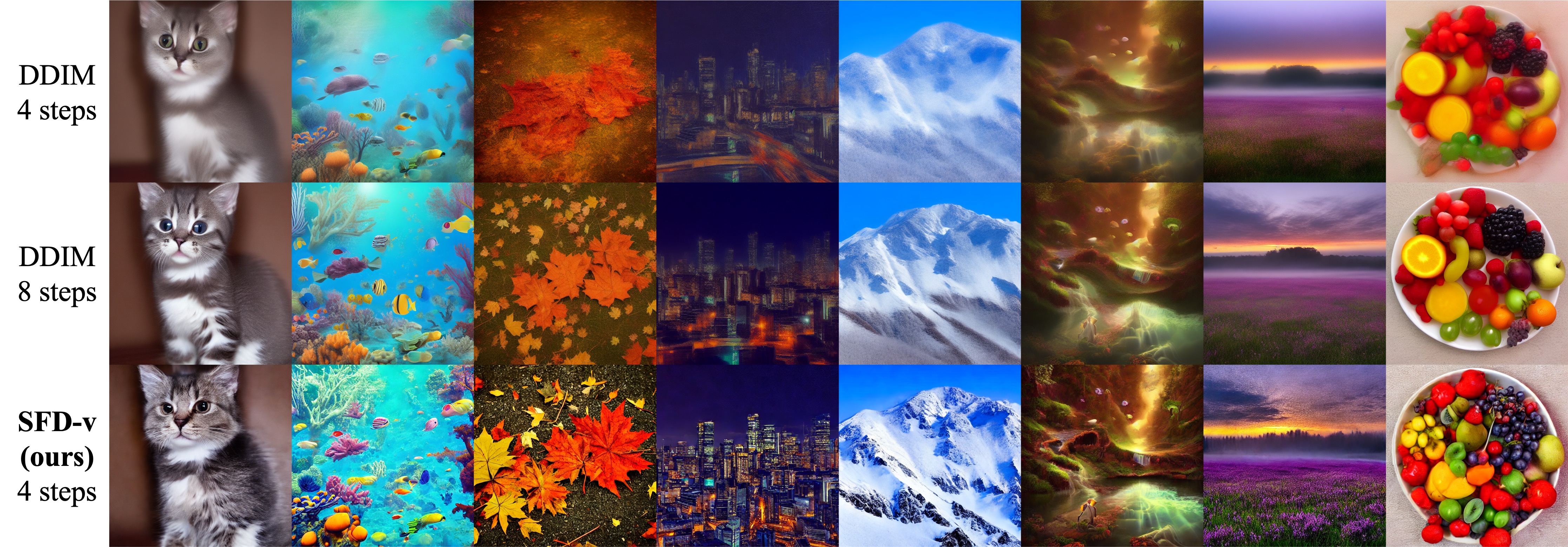}
  \caption{Comparison of synthesized images by Stable Diffusion v1.5~\cite{rombach2022ldm} with guidance scale 7.5.}
  \label{fig:sd}
  \vspace{-\intextsep}
\end{figure}

\begin{compactitem}
    \item 
    \textbf{The mismatch between fine-tuning and sampling steps}. 
    There often exists significant fine-tuning costs in existing distillation-based methods that does not effectively contribute to the final performance due to the step mismatch. 
    For example, progressive distillation~\cite{salimans2022progressive,meng2023distillation,berthelot2023tract} fine-tunes the diffusion model at thousands of timestamps but only a few steps (e.g., 8 or fewer) are used in sampling. Besides, consistency-based distillation~\cite{song2023consistency} spends most fine-tuning efforts to ensure the consistency property~\cite{daras2023consistent}, yet only 1 or 2 steps are used in sampling. Such inconsistencies waste excessive efforts in the fine-tuning process.
    \item 
    \textbf{The complex optimization objectives}. The optimization objectives of distillation-based methods are getting increasingly complex, including the use of LPIPS~\cite{zhang2018unreasonable,song2023consistency,kim2023consistency,yin2023one}, adversarial training~\cite{sauer2023adversarial,kim2023consistency} as well as various regularization terms~\cite{kim2023consistency,yin2023one}. Despite the improved results, these additional components complicate the fine-tuning process.
\end{compactitem}

In this paper, we introduce {\bf S}imple and {\bf F}ast {\bf D}istillation (\ourName) of diffusion models, which aims to achieve fast and high-quality synthesis with diffusion models in a few sampling steps, at minimal fine-tuning cost.
Starting from the general framework behind distillation-based methods, we address the issue of step mismatch by fine-tuning only a small number of timestamps that will be used in sampling, which significantly improves the fine-tuning efficiency.
The effectiveness of this strategy is underpinned by the key observation that, fine-tuning at a specific timestamp can positively impact the gradient direction at other timestamps (Section~\ref{subsec:smooth}).
Our \ourName is then introduced as a simplified paradigm for the distillation of diffusion models, where the student learns to mimic the teacher's sampling trajectory while minimizing accumulated errors. We release the potential of this simple framework by identifying and addressing several small yet vital factors affecting the performance (Section~\ref{subsec:fast}).
Moreover, we propose a variable-NFE version of our method named \ourName-v, which enables a single distilled model to achieve sampling with various steps by introducing a \textit{step-condition} into the model (Section~\ref{subsec:scalable}). 

With 2 NFE, our \ourName achieves a FID of $4.53$ on CIFAR-10~\cite{krizhevsky2009learning} with a training cost of just \textbf{0.64 hours} on a single NVIDIA A100 GPU, which is 1000$\times$ faster than consistency distillation requiring about 1156 hours (see Figure \ref{fig:teaser}). Quantitative and qualitative results on additional datasets, including ImageNet 64$\times$64~\cite{russakovsky2015ImageNet}, Bedroom 256$\times$256~\cite{yu2015lsun}, and image generation with Stable Diffusion~\cite{rombach2022ldm}, demonstrate the effectiveness and efficiency of our methods. 

%% file: sec/preliminary.tex
\section{Preliminary}
\label{sec:preliminary}

\subsection{Diffusion Models}
Diffusion models bridge the implicit data distribution $p_d$ and a Gaussian distribution $p_n$ by progressively adding white Gaussian noise to the data and then iteratively reconstructing the original data from pure noise. 
Diffusion models are grounded in a theoretical framework based on stochastic differential equations (SDEs)~\cite{song2021sde}, with the forward process injecting noise to data:
\begin{equation}
    \label{eq:forward_sde}
    \rmdx = \bff(\bfx, t)\rmd t + g(t) \rmd \bfw_t, 
\end{equation}
where $\bff(\cdot, t): \bbR^{d} \rightarrow \bbR^d, g(\cdot): \bbR \rightarrow \bbR$ are drift and diffusion coefficients, and $\bfw_t \in \bbR^d$ denotes the Wiener process~\cite{oksendal2013stochastic}. 
The backward process reconstructs the original data from the noisy input, which can be achieved with a {\em reverse-time} SDE that shares the same marginals determined by the forward SDE, i.e., 
$\rmdx = [\bff(\bfx, t)- g^2(t)\nablax \log p_{t}(\bfx)  ]\rmd t + g(t) \rmd \bar{\bfw}_t$,
where $\nablax \log p_{t}(\bfx)$ is known as \textit{score function}~\cite{hyvarinen2005estimation,lyu2009interpretation}. The reverse-time SDE can be further simplified to the {\it probability flow ordinary differential equation} (PF-ODE)~\cite{song2021sde,karras2022edm,chen2023geometric}, $\rmdx = [\bff(\bfx, t)- \frac{1}{2}g^2(t)\nablax \log p_{t}(\bfx)]\rmd t$. In particular, we consider $\bff(\bfx, t)=\mathbf{0}$ and $g(t)=\sqrt{2t}$ in this paper, i.e.,
\begin{equation}
    \label{eq:pf_ode}
    \rmdx = -t \nablax \log p_{t}(\bfx) \rmd t, 
\end{equation}
The score function is estimated as $\nablax \log p_{t}(\bfx) \approx -\bfeps_{\theta}(\bfx,t) / t$ with a noise-prediction model $\bfeps_{\theta}(\bfx,t)$ which is obtained by minimizing a regression loss with the weighing function $\lambda(t)$ for each $t$~\cite{ho2020ddpm,song2021ddim,zhou2023fast}:
\begin{equation}
    \label{eq:dm_training}
    \mathcal{L}_{t}(\theta) = \lambda(t) \mathbb{E}_{\bfx \sim p_d, \bfeps \sim \mathcal{N}(\mathbf{0},\bfI)} \lVert \bfeps_{\theta}(\bfx + t\bfeps,t) - \bfeps \rVert_2^2.
\end{equation}
With the noise-prediction model in place of the score function, the PF-ODE can be written as follows
\begin{equation}
    \label{eq:plug-in_ODE}
    \rmdx = \bfeps_{\theta}(\bfx, t)\rmd t.
\end{equation}
Compared to the general reverse-time SDEs, the PF-ODE is preferred in practice for its conceptual simplicity and efficient sampling~\cite{song2021sde,chen2023geometric}.
To sample from a diffusion model with $N$ steps, one first draws $\bfx_N \sim p_n = \calN(\mathbf{0}, t_{\max}^2\bfI)$ and then numerically solves Eq. \ref{eq:plug-in_ODE} by a solver-based method~\cite{song2021ddim,lu2022dpm,lu2022dpmpp,zhang2023deis,zhou2023fast,chen2024trajectory}, following a hand-crafted time schedule $\Gamma(N)=\{t_0=t_{\min}, t_1, \cdots, t_N=t_{\max}\}$. The obtained sample sequence $\{\bfx_{n}\}_{n=0}^N$ is called the \textit{sampling trajectory}.

\subsection{Distillation-based Diffusion Sampling}
Through the lens of Eq. \ref{eq:plug-in_ODE}, we can interpret the noise-prediction model as a gradient field evolves over time, guiding samples towards the data distribution's manifold. Solver-based sampling methods do not change the gradient field and are convenient to implement~\cite{song2021ddim,lu2022dpm,lu2022dpmpp,zhang2023deis,zhou2023fast,chen2024trajectory}. 
However, discretization errors prevent these methods from generating high-quality images within a few sampling steps. Distillation-based methods address this issue by fine-tuning the gradient field with the reference signals provided by a teacher (mostly a solver-based method) to build ``shortcuts'' on the sampling trajectory~\cite{salimans2022progressive,song2023consistency,meng2023distillation,kim2023consistency}. This basic framework behind distillation-based methods, which we call \textit{Trajectory Distillation}, is illustrated in Algorithm \ref{alg:local_td}. 
Specifically, starting from latent encodings $\bfx_{n+1}$ and $\tilde{\bfx}_{n+1}$ ($0 \leq n \leq N-1$) with a sampled $n$, the sampling process is written as:
\begin{align}
    {\rm{Teacher:\quad}} &\tilde{\bfx}_n = {\rm{Solver}}(\tilde{\bfx}_{n+1}, t_{n+1}, t_n, K; \theta),\\
    {\rm{Student:\quad}} &\bfx_n^\psi = {\rm{Euler}}(\bfx_{n+1}, t_{n+1}, t_n, 1; \psi) = \bfx_{n+1} + (t_{n+1} - t_n) \bfeps_\psi(\bfx_{n+1}, t_{n+1}), 
\end{align}
where $K$ is the number of teacher sampling steps taken from $t_{n+1}$ to $t_n$; and $\psi$, $\theta$ are the parameters of the student and teacher model, respectively. 
In each training iteration, the student model $\psi$ is updated with the calculated loss function $\mathcal{L}(\psi) = d(\bfx_n^\psi,\tilde{\bfx}_n)$ using a distance metric $d(\cdot,\cdot)$. ${\rm{Solver}}(\cdot,\cdot,\cdot,\cdot;\theta)$ can be any solver-based method with the fixed $\theta$ to provide reference signals. For example, in progressive distillation~\cite{salimans2022progressive,meng2023distillation}, it is defined as the Euler sampler~\cite{song2021ddim} with $K=2$, while in consistency distillation~\cite{song2023consistency,kim2023consistency}, the Heun sampler~\cite{karras2022edm}, $K=1$, and a consistency loss are used. 

Distillation-based methods have demonstrated impressive results but generally incur a significant computational overhead. In the following sections, we revisit the trajectory distillation framework and unlock its potential through a comprehensive assessment of the key factors affecting the performance. 

\vspace{-1.5\baselineskip}
\begin{minipage}[t]{0.49\textwidth}
\begin{algorithm}[H]
\caption{Trajectory Distillation}
  \begin{algorithmic}
  \label{alg:local_td}
  \REPEAT
    \STATE Sample $\bfx_0$ from the dataset
    \STATE Sample $n \sim \mathcal{U}(0,N-1)$
    \STATE Sample $\bfx_{n+1} \sim \mathcal{N}(\bfx_0;{t_{n+1}^2}\bfI)$
    \STATE $\bfx_n^\psi \leftarrow {\rm{Euler}}(\bfx_{n+1}, t_{n+1}, t_n, 1; \psi)$
    \STATE $\tilde{\bfx}_n \leftarrow {\rm{Solver}}(\bfx_{n+1}, t_{n+1}, t_n, K; \theta)$
    \STATE $\mathcal{L}(\psi) \leftarrow d(\bfx_n^\psi, \tilde{\bfx}_n)$
    \STATE $\psi \leftarrow \psi - \eta\nabla_\psi\mathcal{L}(\psi)$
  \UNTIL convergence
  \end{algorithmic}
\end{algorithm}
\end{minipage}
\hfill
\begin{minipage}[t]{0.5\textwidth}
\begin{algorithm}[H]
\caption{SFD (ours)}
  \begin{algorithmic}
  \label{alg:global_td}
  \REPEAT
    \STATE Sample $\bfx_{N} = \tilde{\bfx}_{N} \sim \mathcal{N}(\mathbf{0};{t_{N}^2}\bfI)$
    \FOR{$n=N-1$ \textbf{to} $0$}
      \STATE $\bfx_n^\psi \leftarrow {\rm{Euler}}(\bfx_{n+1}, t_{n+1}, t_n, 1; \psi)$
      \STATE $\tilde{\bfx}_n \leftarrow {\rm{Solver}}(\tilde{\bfx}_{n+1}, t_{n+1}, t_n, K; \theta)$
      \STATE $\psi \leftarrow \psi - \eta\nabla_\psi d(\bfx_n^\psi, \tilde{\bfx}_n)$
      \STATE $\bfx_n \leftarrow {\rm{detach}}(\bfx_n^\psi)$
    \ENDFOR
  \UNTIL convergence
  \end{algorithmic}
\end{algorithm}
\end{minipage}

%% file: sec/method.tex
\section{Method}
\label{sec:method}

\subsection{Smooth Modification of the Gradient Field}
\label{subsec:smooth}

\begin{wrapfigure}{r}{0.35\textwidth}  
  \vspace{-\intextsep}
  \includegraphics[width=0.35\textwidth]{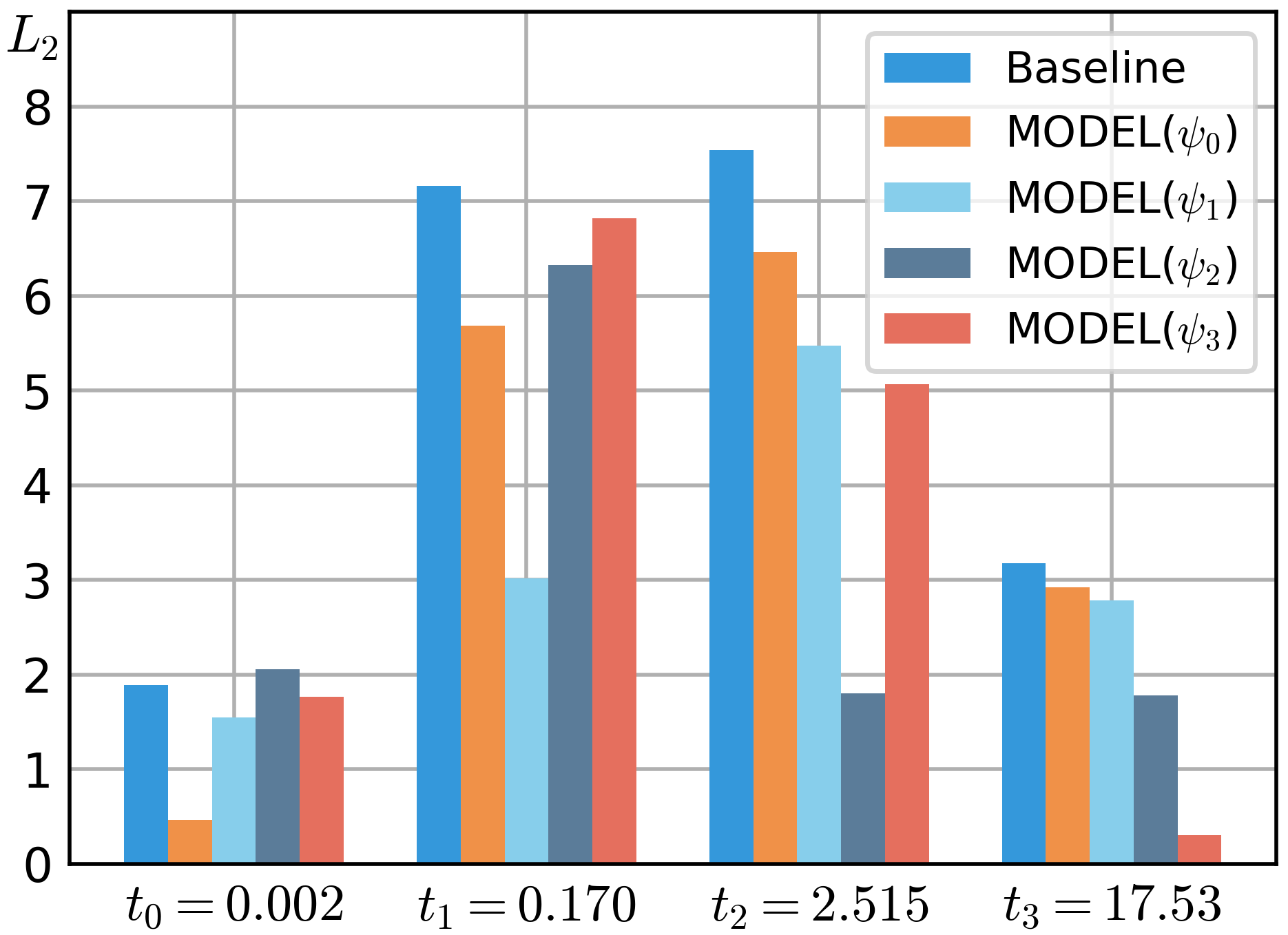}
  \caption{\em \small MODEL($\psi_n$) is trained to match the teacher's sampling trajectory at $t_n$ but can enhance the matching at untrained timestamps. The time schedule follows the polynomial schedule with $\rho=7, t_0=0.002, t_4=80$.}
  \label{fig:observation}
  \vspace{-\intextsep}
\end{wrapfigure}

As mentioned in Section \ref{sec:intro}, existing distillation-based methods incur significant fine-tuning costs that may not effectively contribute to the final sample quality~\cite{salimans2022progressive,meng2023distillation,song2023consistency}, which is a key factor overburdening computational resources. Instead, we propose to fine-tune only a few timestamps that will be used in sampling.
To validate our strategy, we initialize four different student models (denoted as MODEL($\psi_n$), $0 \leq n \leq N-1$) from a pre-trained teacher model $\theta$ using the second-order DPM-Solver(2S)~\cite{lu2022dpm} with $N=4$ and $K=3$.
We then fine-tune each MODEL($\psi_n$) only on a certain timestamp $t_{n+1}$ and make it align with the teacher predictions at the next timestamp $t_n$. After fine-tuning, we evaluate the performance of each MODEL($\psi_n$) at all timestamps by comparing with the teacher's sampling trajectory $\{x_{n}\}_{n=0}^N$ under the same setting.
Specifically, we calculate the $L_2$ distance in the following two formulas for all $0 \leq n,k \leq N-1$, 
\begin{align}
    {\rm{Baseline:}}\quad &\lVert x_n - {\rm{Euler}}(\bfx_{n+1}, t_{n+1}, t_n, 1; \theta) \rVert_2, \\
    {\rm{MODEL}}(\psi_k):\quad &\lVert x_n - {\rm{Euler}}(\bfx_{n+1}, t_{n+1}, t_n, 1; \psi_k) \rVert_2,
\end{align}
and average the results over 1,000 trajectories. As shown in Figure \ref{fig:observation}, the distance calculated using the fine-tuned models is almost consistently smaller than that of the baseline. This is remarkable since each MODEL($\psi_n$) is only trained to match the teacher's sampling trajectory at a specific $t_n$. Yet, its performance on other timestamps is mostly improved, even though different timestamps are far apart. This indicates that trajectory distillation does not disrupt the gradient field but enhances it smoothly. Since fine-tuning at different timestamps mutually reinforces the model, fine-tuning on a fine-grained time schedule is unnecessary. This insight forms the basis of our strategy. Beyond efficiency, we will demonstrate that our approach achieves high performance in the sequel.

\subsection{Simple and Fast Distillation of Diffusion Models}
\label{subsec:fast}

\begin{figure}[t]
  \centering
  \begin{subfigure}[b]{0.32\textwidth}
    \includegraphics[width=\textwidth]{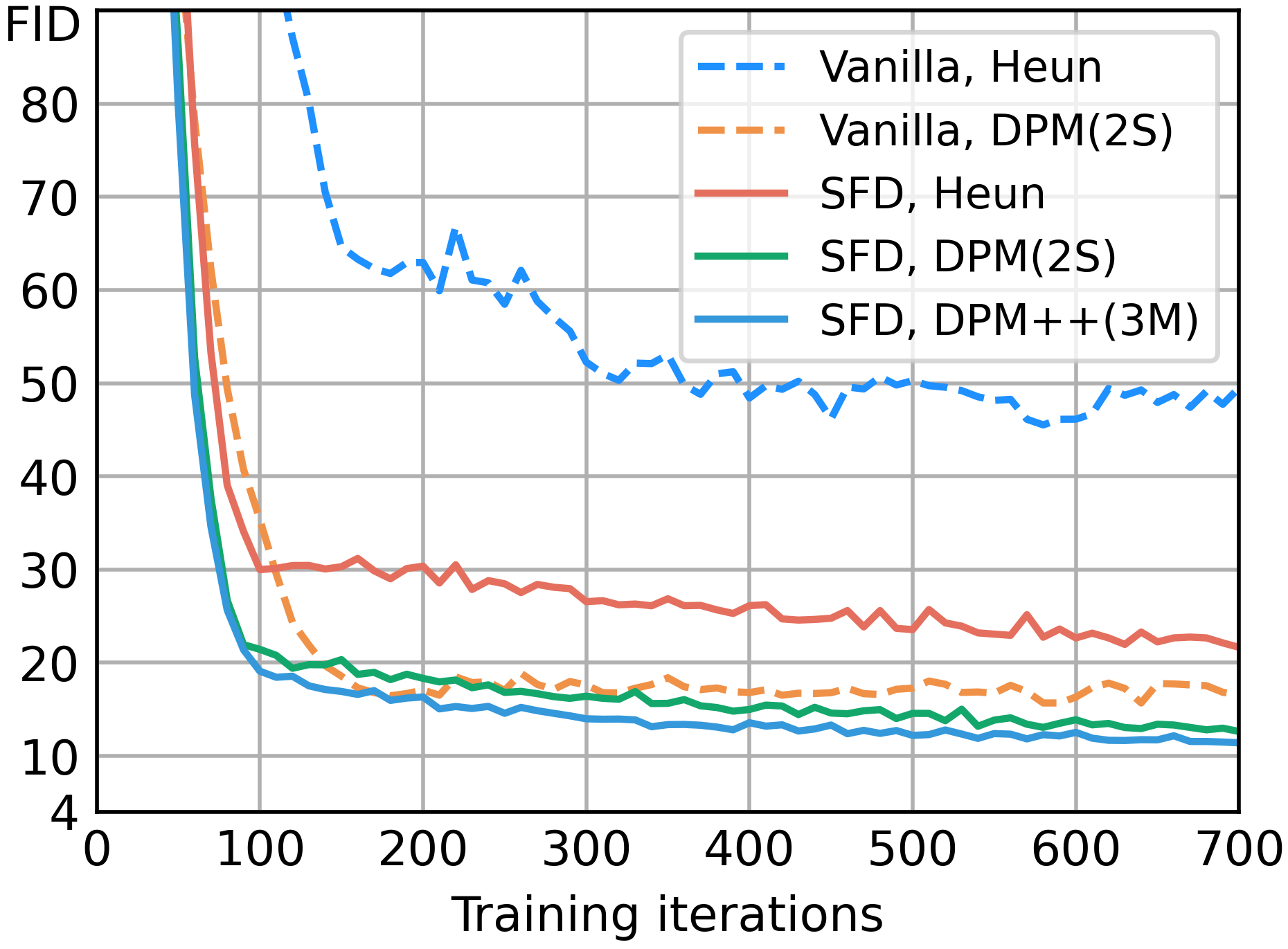}
    \caption{Teacher solver.}
    \label{fig:ablation_solver}
  \end{subfigure}
  \begin{subfigure}[b]{0.32\textwidth}
    \includegraphics[width=\textwidth]{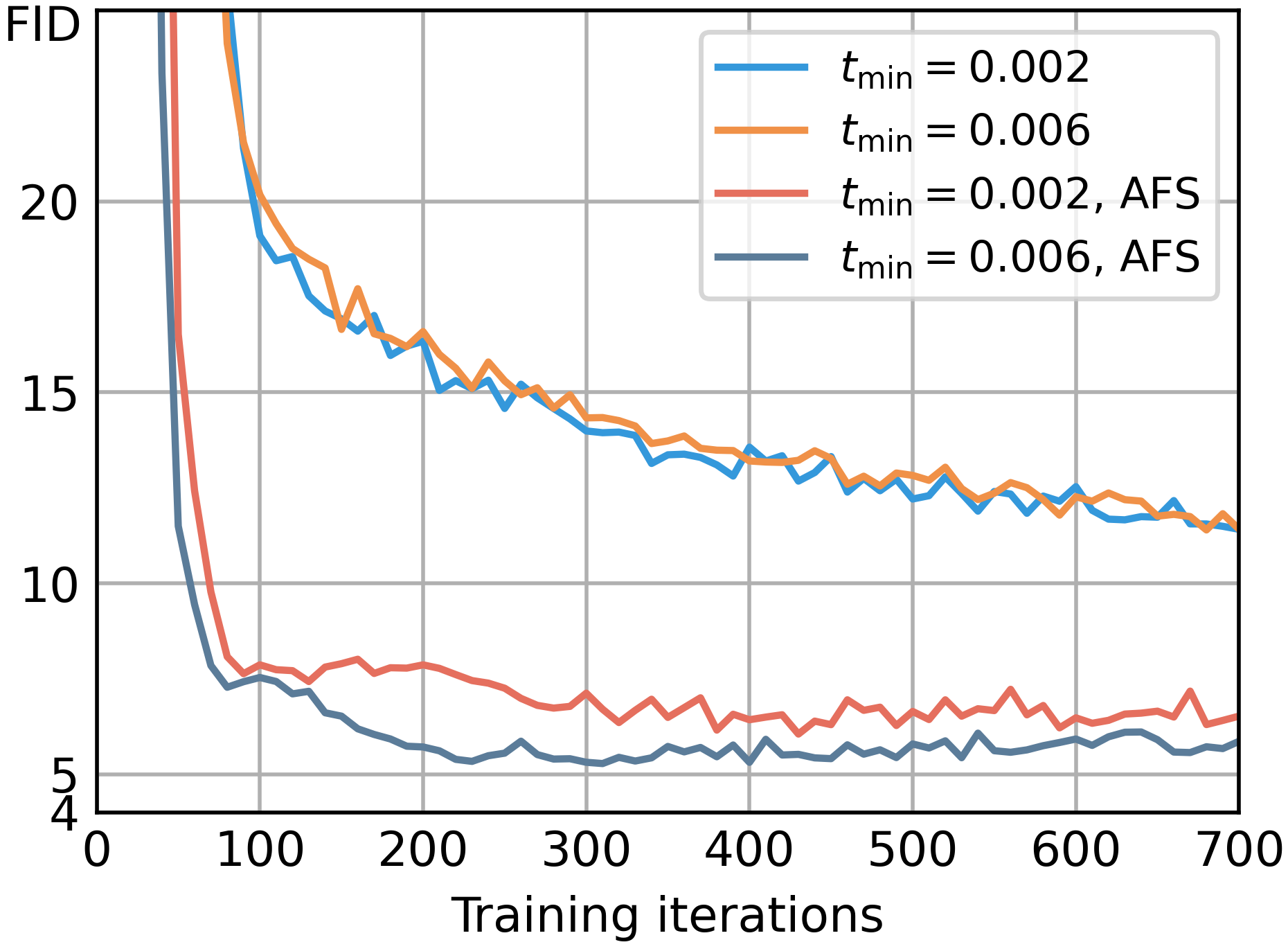}
    \caption{Timestamps.}
    \label{fig:ablation_time}
  \end{subfigure}
  \begin{subfigure}[b]{0.32\textwidth}
    \includegraphics[width=\textwidth]{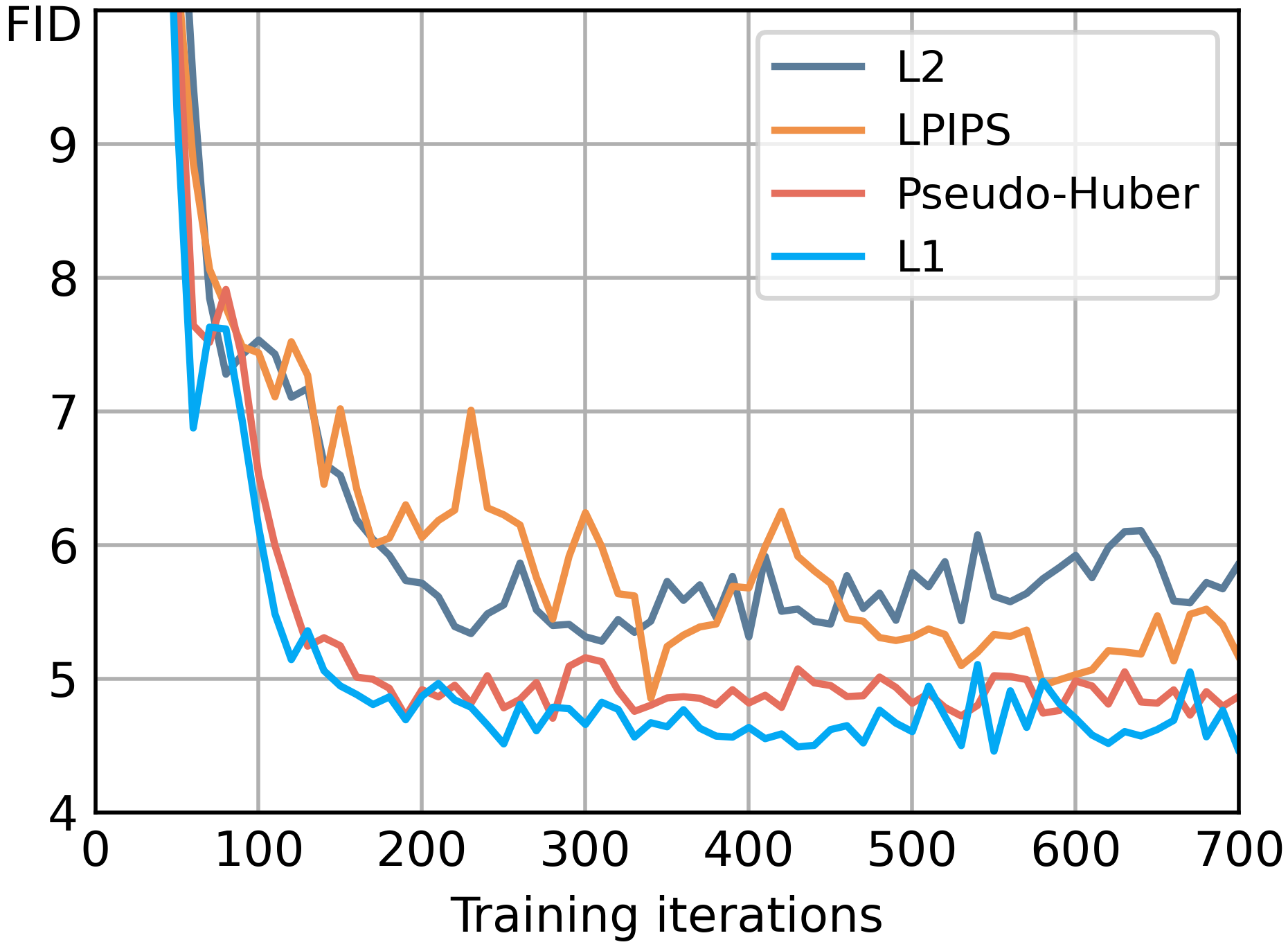}
    \caption{Loss metric.}
    \label{fig:ablation_loss}
  \end{subfigure}
  \caption{Ablation studies of 2-NFE distillation on CIFAR10. The FID is evaluated by 50,000 generated samples with the same latent encodings and is reported every 10 iterations. 
  We achieve the best performance with SFD, DPM-Solver++(3M) teacher, AFS, $t_{\min}=0.006$ and L1 loss.}
\end{figure}

As for solver-based methods, the cost of a single sampling step varies depending on the design, which is commonly measured by the number of function evaluations (NFE). 
For the DDIM sampler~\cite{song2021ddim} and other higher-order methods such as DPM-Solver++(3M)~\cite{lu2022dpmpp} and UniPC~\cite{zhao2023unipc}, one sampling step corresponds to one NFE, while two NFEs are required for the Heun sampler~\cite{karras2022edm} and DPM-Solver(2S)~\cite{lu2022dpm}. In the following, we distill a diffusion model to achieve sampling with two NFEs ($N=2$ by default). We configure a reasonable baseline on the CIFAR10 dataset~\cite{krizhevsky2009learning} and gradually improve the performance through extensive experiments. 
The improved configuration is proven to be effective across different NFEs and datasets.

\noindent
\textbf{Default configuration}. The Heun sampler is used to generate the teacher sampling trajectory instead of DDIM for efficiency, which has been demonstrated in training consistency models~\cite{song2023consistency,kim2023consistency}. We set $K=3$ for Heun sampler, which gives sampling trajectories with 12 NFEs. For the time schedule, if not otherwise specified, we use the polynomial schedule where $\rho=7$, $t_{\min}=0.002$ and $t_{\max}=80$, following the EDM implementation~\cite{karras2022edm,song2023consistency}. The squared L2 loss is used by default. 
For experiments in this section, we use a batch size of 128, learning rate of 5e-5 and fine-tune with 100,000 teacher sampling trajectories generated (around 780 training iterations).

\noindent
\textbf{From local to global}.
We start by analyzing the potential defects of trajectory distillation. As shown in Algorithm \ref{alg:local_td}, trajectory distillation performs local fine-tuning. The term ``local'' indicates that the teacher model only generates part of the sampling trajectory (i.e., from $t_{n+1}$ to $t_n$), and the optimization is independent across different $n$. This raises two defects that limit both efficiency and performance: 
(i) Higher-order multi-step solvers that require history evaluation records are unsupported. 
(ii) The student model is trained to imitate only part of the teacher's sampling trajectory. During sampling, the errors accumulate since the student is never trained to perfectly fix them.
To address these issues, we view trajectory distillation from a global perspective and introduce our {\em {\bf S}imple and {\bf F}ast {\bf D}istillation} of diffusion models (\ourName) in Algorithm \ref{alg:global_td}. In each training iteration of \ourName, we first generate the whole teacher sampling trajectory and let the student imitate it step by step. During this process, the student model generates its own trajectories, enabling it to learn to fix the accumulated errors.
In Figure \ref{fig:ablation_solver}, we compare both strategies (marked as ``Vanilla'' and ``SFD'') using the default configuration. It is shown that \ourName exhibits better performance. In the following sections, we focus on \ourName and seek to release its potential for efficient distillation of diffusion models.

\noindent
\textbf{Efficient solver}. One of the key components that affects the efficiency of distillation-based methods is the choice of the teacher solver. To compare the performance of different solvers, we conduct experiments on both trajectory distillation and \ourName with 3 representative solver-based methods: second-order Heun sampler, second-order DPM-Sovler(2S) and third-order DPM-Solver++(3M). Since history evaluations are unavailable, we exclude DPM-Solver++(3M) for trajectory distillation. The NFE of teacher sampling trajectories is kept 12 consistently ($K=6$ is hence used for DPM-Solver++(3M)) and the results are shown in Figure \ref{fig:ablation_solver}. DPM-Solver++(3M) stands out, and the Heun sampler is shown to be suboptimal.
Therefore, for distillation-based methods with trajectory distillation involved, it is recommended to explore replacing the Heun sampler (or DDIM sampler) with DPM-Solver(2S). 
We leave this to future works.

\begin{figure}[t]
  \begin{minipage}[b]{0.31\textwidth}
    \includegraphics[width=\textwidth,bb=1 2 241 195]{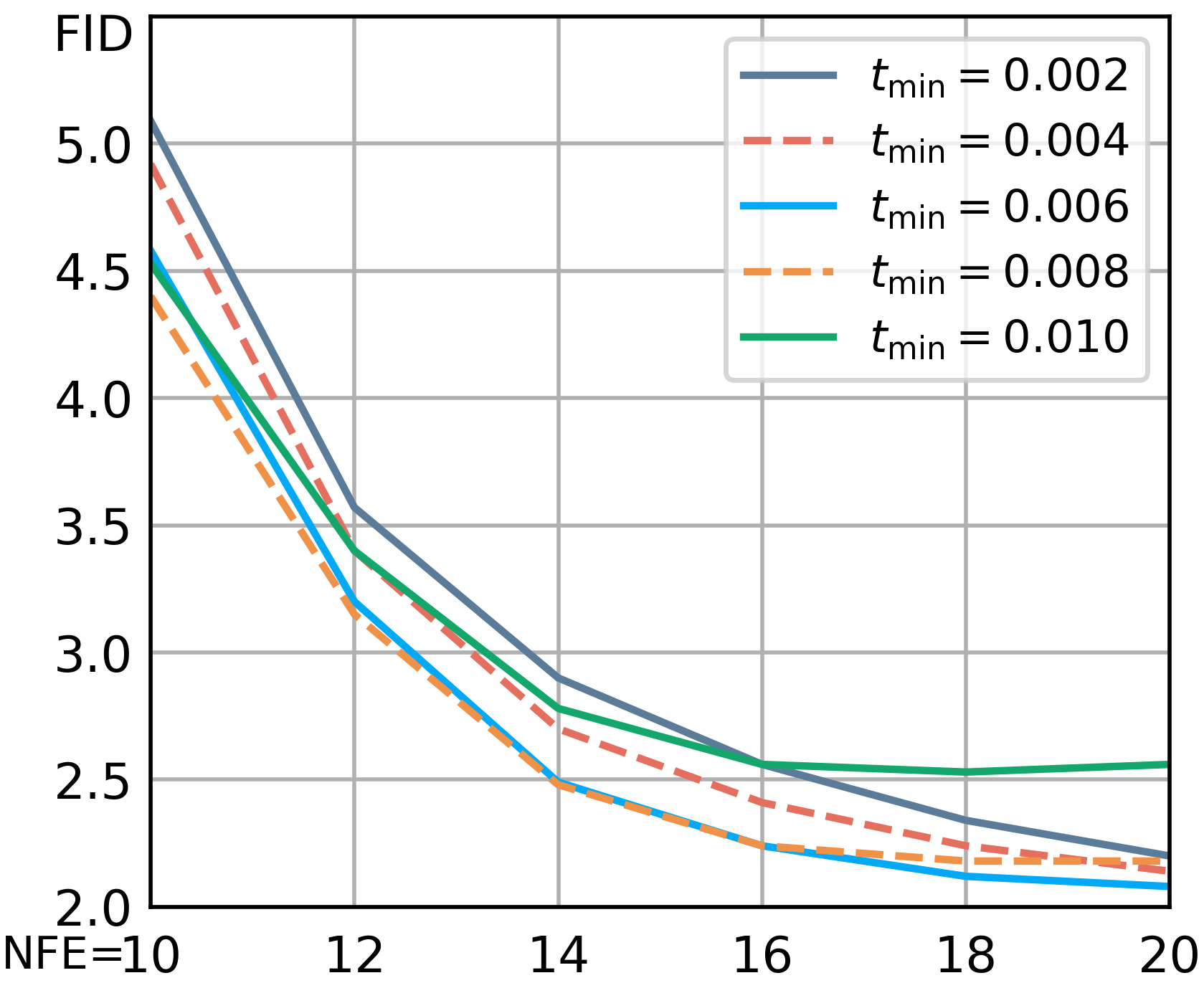}
    \caption{Ablation study on $t_{\min}$ with DPM++(3M).}
    \label{fig:comp_sigma}
  \end{minipage}
  \hfill
  \begin{minipage}[b]{0.31\textwidth}
    \centering
    \includegraphics[width=\textwidth,bb=1 2 256 208]{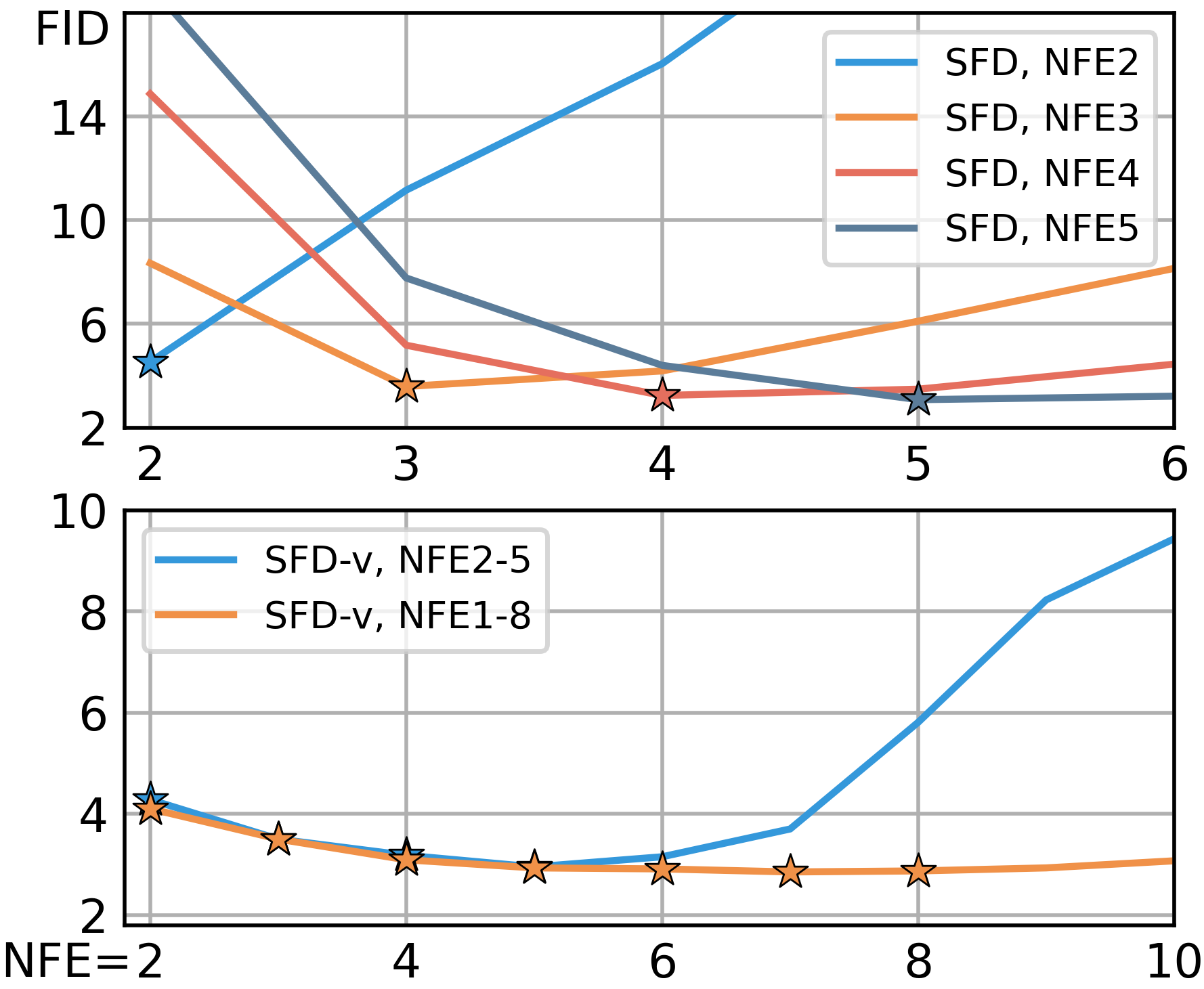}
    \caption{Extrapolation ability on untrained NFE.}
    \label{fig:extrap}
  \end{minipage}
  \hfill
  \begin{minipage}[b]{0.36\textwidth}
    \centering
    \includegraphics[width=\textwidth,bb=4 3 390 281]{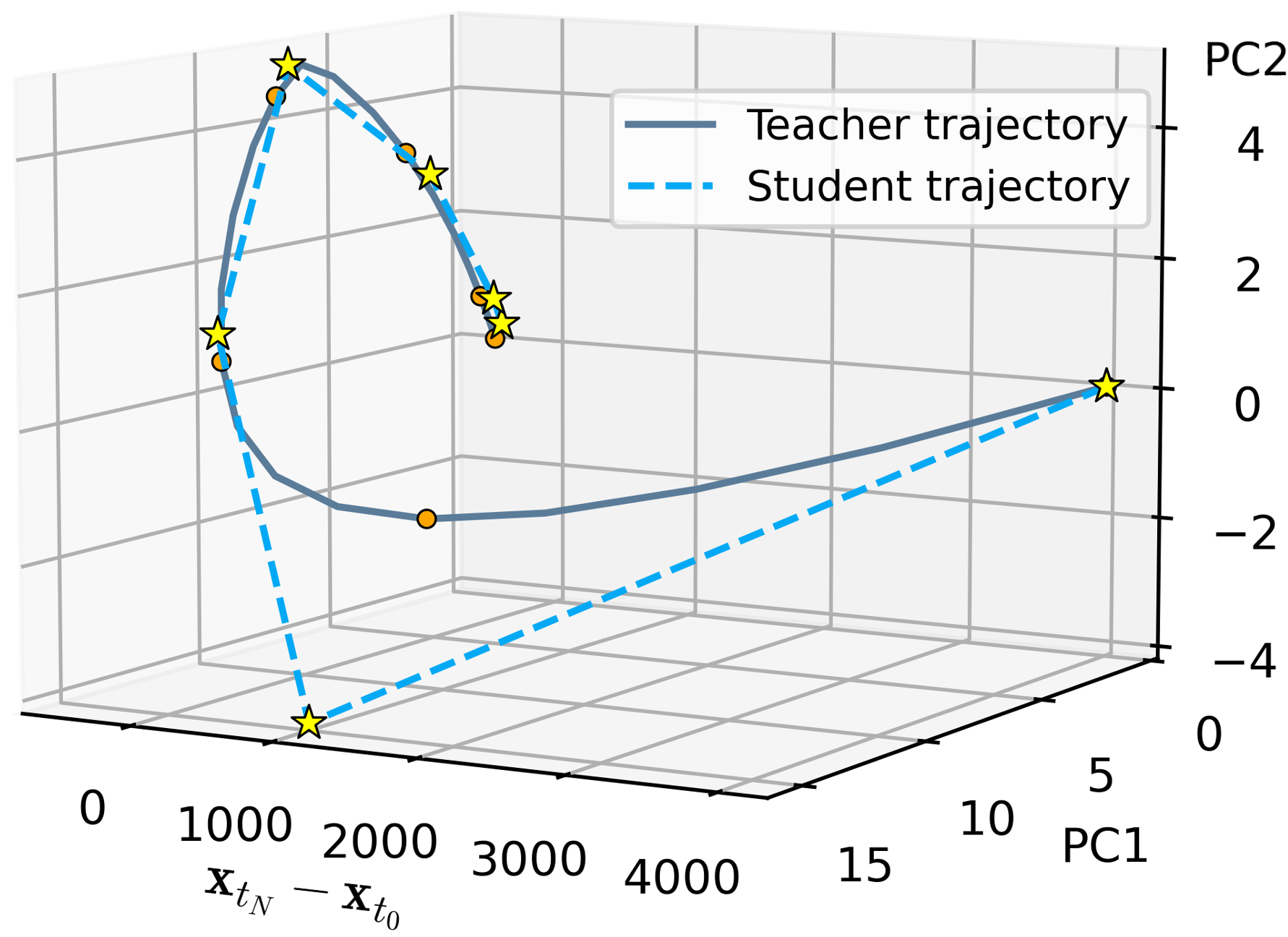}
    \caption{Visualization of the effectiveness of \ourName.}
    \label{fig:distilled_traj}
  \end{minipage}
\end{figure}

\noindent
\textbf{Minimum and maximum timestamps}. Choosing DPM-Solver++(3M) as the teacher solver, we improve \ourName by adjusting the start and end timestamps during training and sampling. For the minimum timestamp $t_{\min}$, we empirically find that a slight increase improves the student and teacher sampling performance across various datasets. An ablation study of $t_{\min}$ on CIFAR10 dataset is shown in Figure \ref{fig:comp_sigma}. 
We increase $t_{\min}$ from 0.002 to 0.006. This change provides consistent improvements across different pre-trained diffusion models. For the maximum timestamp, we introduce analytical first step (AFS)~\cite{dockhorn2022genie,zhou2023fast,chen2024trajectory} in the generation of student sampling trajectories, which takes one estimated step $\bfeps_{\psi}(\bfx_N, t_N) \approx \bfx_N / \sqrt{1 + t_N^2}$ at the beginning of sampling to save one NFE. 
Therefore, to obtain a 2-NFE \ourName with AFS applied, we use $N=3$ and $K=4$. As shown in Figure \ref{fig:ablation_time}, using AFS largely boosts the performance of \ourName, indicating that one more inaccurate step can outperform one less step. This improvement also benefits from the nature of \ourName, where the error incurred in the first step can be fixed by later steps (see the visualization in Figure \ref{fig:visualization}). The detailed algorithm of \ourName with AFS is included in Appendix \ref{sec:algo}. 

\noindent
\textbf{Loss metric}. In Figure \ref{fig:ablation_loss}, we test various distance metrics for the loss function including squared L2 distance, L1 distance, LPIPS distance~\cite{zhang2018unreasonable} and Pseudo-Huber distance~\cite{song2023improved}. Among these metrics, L1 distance outperforms. Note that the LPIPS distance is trained to evaluate the perceptual distance between two images but not corrupted ones, which may explain its suboptimal performance.

\begin{wraptable}{r}{0.5\textwidth}
  \vspace{-\intextsep}    
  \caption{Quantitative results of the ablations.}
  \label{tab:quanti_ablation}
  \centering
  \fontsize{7}{9}\selectfont
  \begin{tabular}{lccccr}
    \toprule
    Method & Teacher & $t_{\min}$ & AFS & Loss & FID \\
    \midrule
    Vanilla & Heun       & 0.002 & N/A   & L2 & 46.84 \\
    Vanilla & DPM(2S)    & 0.002 & N/A   & L2 & 16.69 \\
    SFD     & Heun       & 0.002 & False & L2 & 20.88 \\
    SFD     & DPM(2S)    & 0.002 & False & L2 & 12.50 \\
    SFD     & DPM++(3M)  & 0.002 & False & L2 & 11.65 \\
    SFD     & DPM++(3M)  & 0.006 & False & L2 & 10.93 \\
    SFD     & DPM++(3M)  & 0.002 & True  & L2 & 7.17  \\
    SFD     & DPM++(3M)  & 0.006 & True  & L2 & 5.67  \\
    SFD     & DPM++(3M)  & 0.006 & True  & LPIPS & 5.10  \\
    SFD     & DPM++(3M)  & 0.006 & True  & PH & 4.90  \\
    SFD     & DPM++(3M)  & 0.006 & True  & L1 & 4.57  \\
    \bottomrule
  \end{tabular}
  \vspace{-\intextsep}    
\end{wraptable}

With these improvements, the \ourName achieves a fast convergence with only around 300 training iterations, which only takes around \textbf{8 minutes} on a single NVIDIA A100 GPU. The performance of the obtained 2-NFE \ourName is even comparable with the 2-NFE model trained by progressive distillation~\cite{salimans2022progressive,meng2023distillation}, which takes more than 100 hours under our estimation. 
The quantitative results are included in Table \ref{tab:quanti_ablation}.

To verify our findings in Section \ref{subsec:smooth}, we test the extrapolation ability of \ourName with untrained NFEs on CIFAR10. The results are shown in Figure \ref{fig:extrap} where the markers indicate the NFEs our methods are trained to sample with.
Take ``\ourName, NFE4'' as an example, where the \ourName is only trained to sample with NFE of 4 and its performance is marked by a star. When using this \ourName to sample with untrained NFEs (i.e., 2,3,5,6), even though the timestamps have never been trained in these cases, the performance is still decent and largely outperforms the DDIM sampler (DDIM with NFE of 6 gives a FID of 35.62, far exceeds the range of the figure).
This empirically verifies our hypothesis that the gradient field is not disrupted but is smoothly enhanced. The change of the gradient field of a certain timestamp can also change that of nearby timestamps in a similar way.

Moreover, in Figure \ref{fig:distilled_traj}, we leverage the three-dimensional projection technique proposed in~\cite{chen2024trajectory} to visualize the sampling trajectories generated by \ourName with 5 NFEs and that of the teacher solver \ourName is trained to imitate. Due to the use of AFS, the first sampling step of \ourName is inaccurate. But the accumulated errors are largely reduced in the following steps thanks to the global distillation used in our \ourName as discussed in Section~\ref{subsec:fast}. We include more visualised trajectories in Appendix \ref{subsec:quanti_and_quali}.

\begin{wrapfigure}{r}{0.3\textwidth}
  \vspace{-1\intextsep}    
  \includegraphics[width=0.3\textwidth]{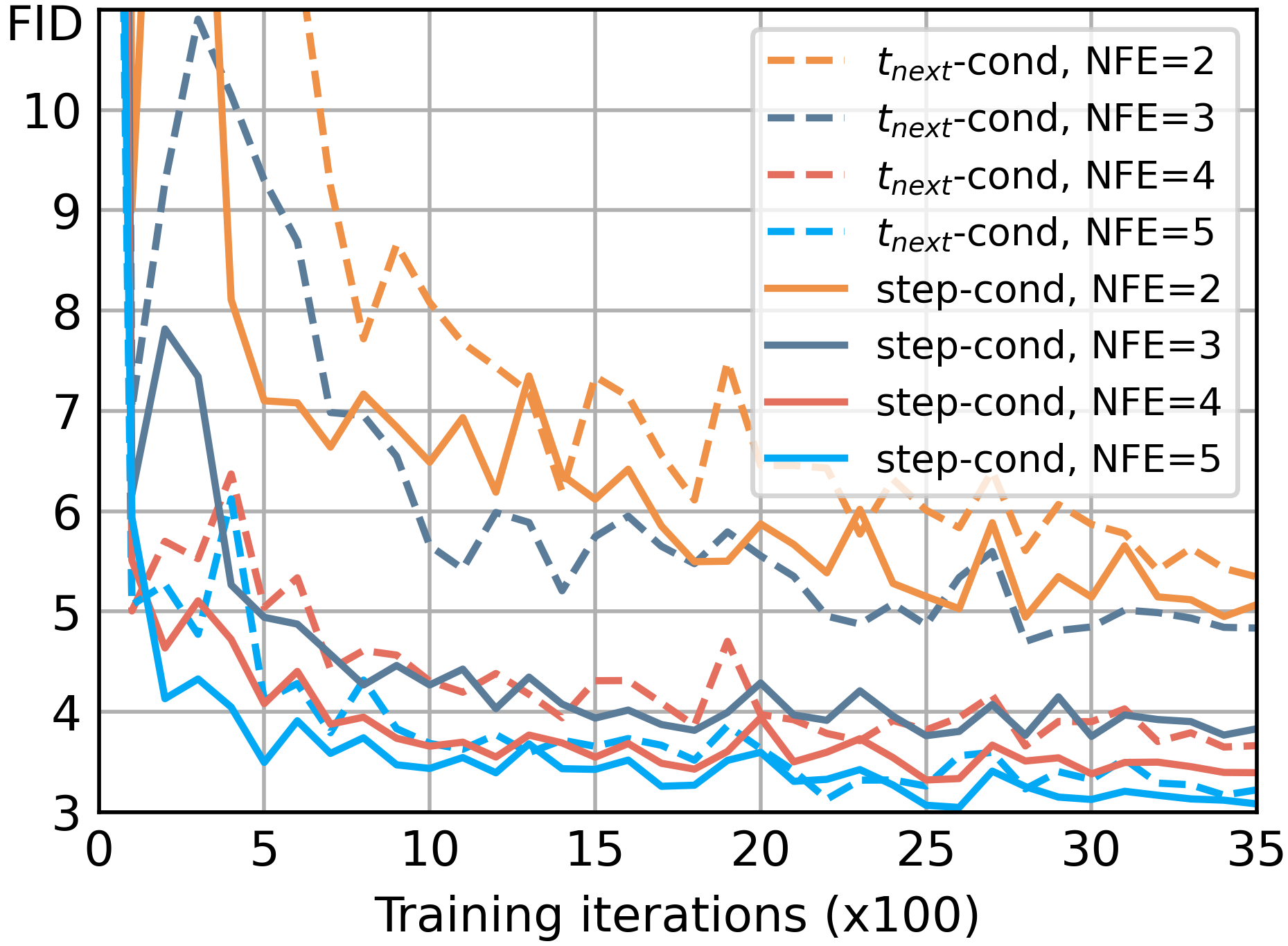}
  \caption{\em \small Ablation study on the type of condition.}
  \label{fig:ablation_cond}
  \vspace{-\intextsep}  
\end{wrapfigure}

\subsection{Towards Variable-NFE Distillation}
\label{subsec:scalable}
One attractive property of diffusion models is that the sample quality can be consistently improved by increasing the sampling steps, which is currently unsupported by most distillation-based methods. Progressive distillation~\cite{salimans2022progressive,meng2023distillation} partially addresses this issue using the multi-stage training. However, the model saved in each training round only supports sampling with a certain trained step (i.e., 1, 2, 4, 8, $\cdots$). The sample quality of consistency models~\cite{song2023consistency,song2023improved} designed for the one-step sampling also deteriorates under larger sampling steps as revealed by~\cite{kim2023consistency}. Moreover, the unique encoding property of ODE-based diffusion sampling is corrupted in multi-step consistency models due to the noise injected in every step.

To address this issue, consistency trajectory models (CTM)~\cite{kim2023consistency} introduce a new condition into the student model, which specifies the next time to arrive, referred to as the $t_{\text{next}}$-condition. 
Unlike CTM, introducing a \textit{step-condition} to our \ourName is more efficient. By informing the student model of the number of sampling steps, our \ourName can perform sampling with different NFEs. We refer to this variable-NFE version of our method as \ourName-v.

In every training iteration of \ourName-v, the total number of sampling steps $N$ is first sampled uniformly from a pre-specified list of steps. Then, the time schedule $\Gamma(N)$ is generated, and the subsequent training process is the same as training \ourName.
As shown in the ablation study in Figure \ref{fig:ablation_cond}, the step-condition consistently outperforms the $t_{next}$-condition. 
For the injection of step-condition, we treat it the same way as the time embedding in diffusion models (see Appendix \ref{subsec:step_cond} for more details). 
We include the algorithm of training \ourName-v in Appendix \ref{sec:algo}.

\subsection{Distillation under Classifier-free Guidance}
\label{subsec:guided_gtd}

Stable Diffusion~\cite{rombach2022ldm}, a latent diffusion model combined with classifier-free guidance~\cite{ho2022classifier}, has shown to be highly effective in high-resolution image generation. 
The classifier-free guidance extends the flexibility of the generation of diffusion models by introducing the guidance scale $\omega$. Given a conditioning information $c$, the noise-prediction model is rewritten as 
\begin{equation}
    \label{eq:cfg}
    \tilde{\bfeps}_{\theta} (\bfx,t,c) = \omega \bfeps_{\theta} (\bfx,t,c) + (1 - \omega) \bfeps_{\theta} (\bfx,t,c=\varnothing).
\end{equation}
However, Stable Diffusion requires a large number of network parameters and sampling steps to produce a satisfying generation. 
Moreover, the cost of every sampling step doubles since both conditional and unconditional evaluations are involved in Eq. \ref{eq:cfg}.
Distilling the Stable Diffusion model into a few steps is challenging because of source-intensive requirements and the flexibility given by the guidance scale. To address this issue, existing methods either introduce an $\omega$-condition into their model~\cite{meng2023distillation,li2024snapfusion,luo2023latent}, or simply discard the guidance scale~\cite{xu2023ufogen,sauer2023adversarial}.

\begin{figure}[t]
  \centering
  \begin{subfigure}[b]{0.32\textwidth}
    \includegraphics[width=\textwidth]{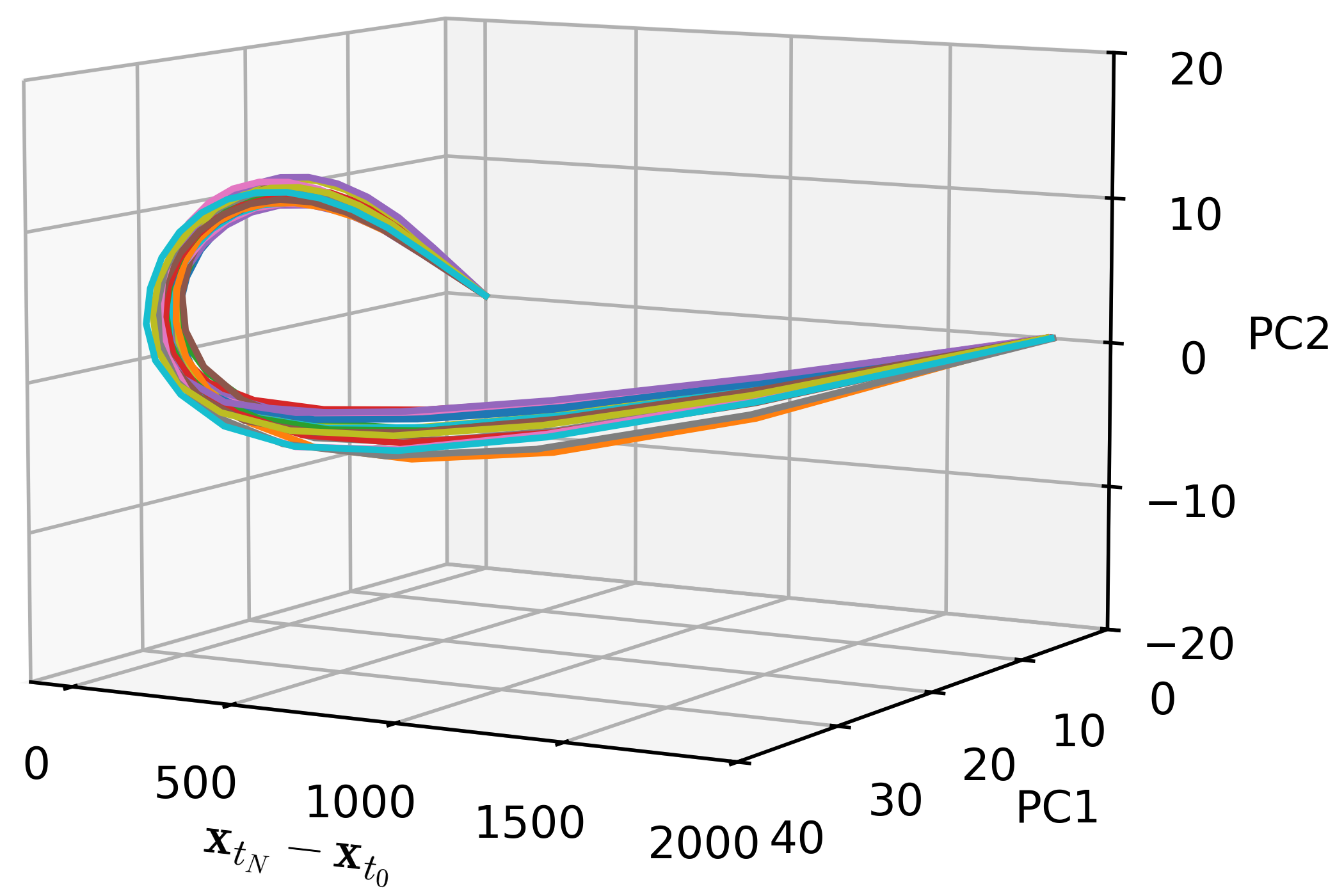}
    \caption{Guidance scale = 1.0.}
    \label{fig:scale1}
  \end{subfigure}
  \begin{subfigure}[b]{0.32\textwidth}
    \includegraphics[width=\textwidth]{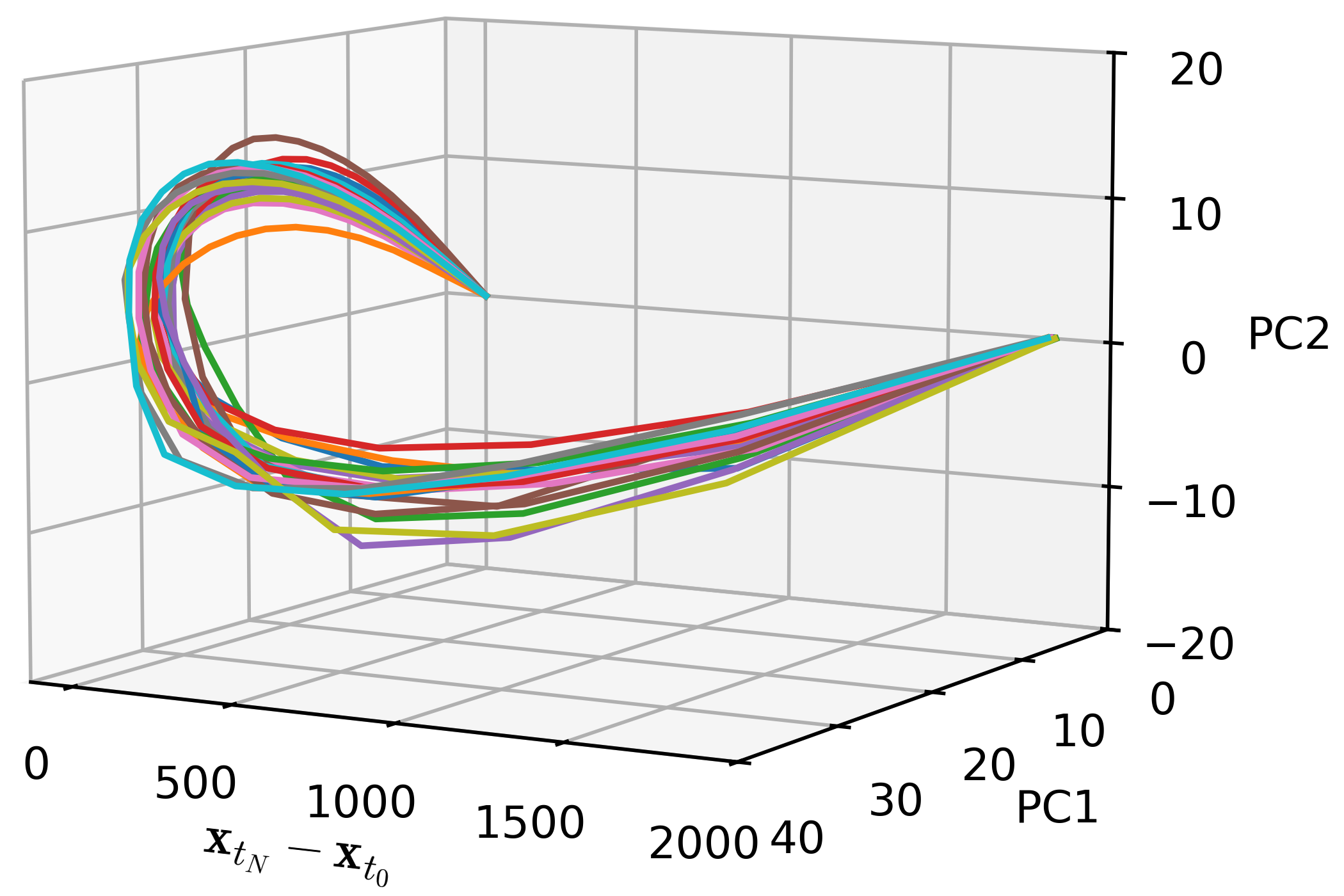}
    \caption{Guidance scale = 4.0.}
    \label{fig:scale4}
  \end{subfigure}
  \begin{subfigure}[b]{0.32\textwidth}
    \includegraphics[width=\textwidth]{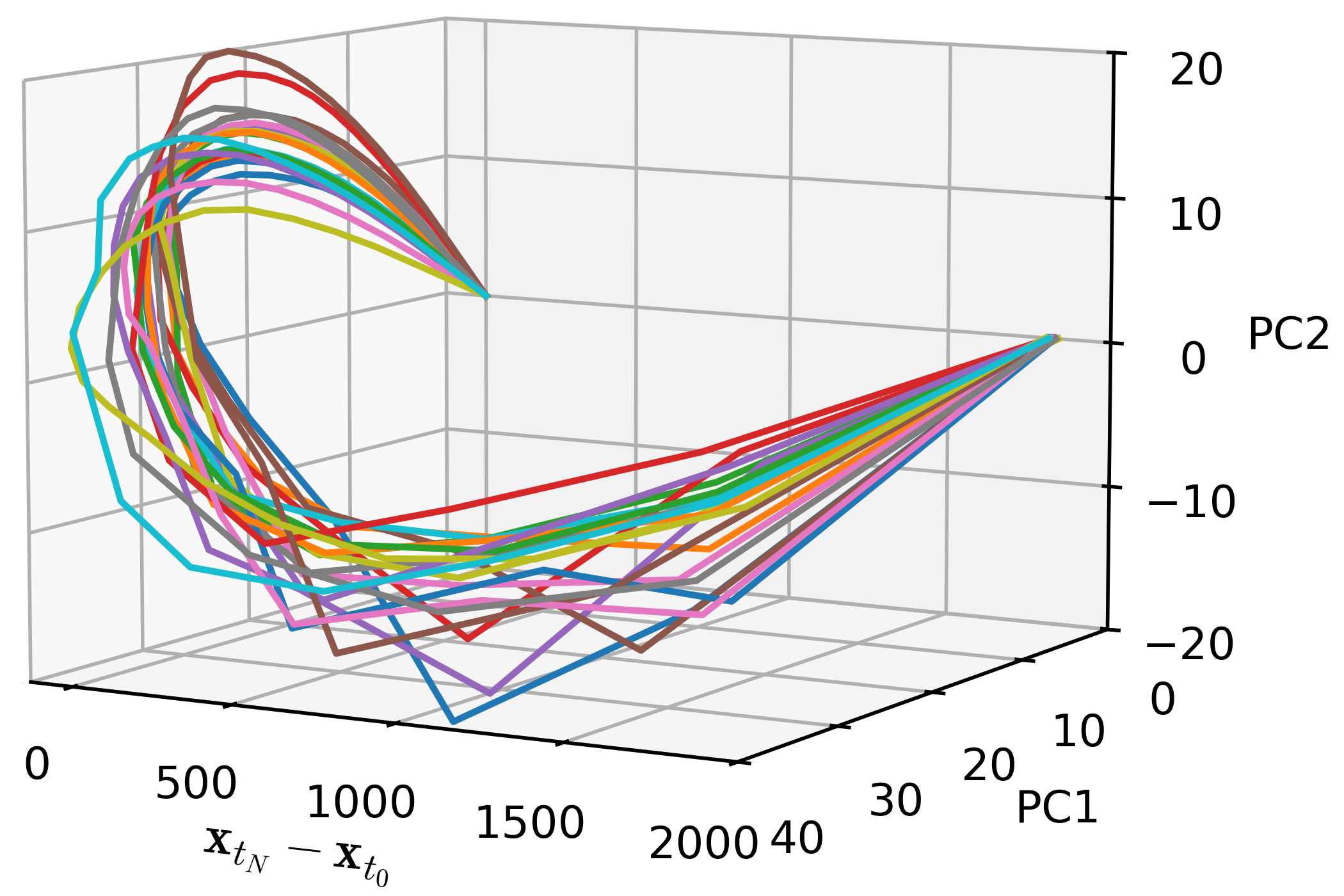}
    \caption{Guidance scale = 7.5.}
    \label{fig:salce7}
  \end{subfigure}
  \caption{We visualize 20 sampling trajectories generated by DPM-Solver++(2M)~\cite{lu2022dpmpp} with 20 steps using the three-dimensional projection technique proposed in~\cite{chen2024trajectory}.}
    \label{fig:visualization}
    \vspace{-0.5\intextsep}
\end{figure}

Here, we propose a new strategy with the observation on the sampling trajectories generated by Stable Diffusion under different guidance scales.
Following the three-dimensional projection technique proposed in~\cite{chen2024trajectory}, we visualize the sampling trajectories generated by Stable Diffusion in its latent space using DPM-Solver++(2M) starting from 20 fixed latent encodings. 
As shown in Figure \ref{fig:visualization}, sampling trajectories projected to the three-dimensional subspace exhibit a regular boomerang shape, which is consistent with the findings in the previous work~\cite{chen2024trajectory}. Furthermore, we observe that the sampling trajectories become more complex as the guidance scale increases, making trajectory distillation on the high guidance scale even more challenging. 
This observation naturally leads to our strategy: \textit{perform distillation with a guidance scale of 1 and sampling with any guidance scale}. Our strategy enables accelerated training since the unconditional evaluation in Eq. \ref{eq:cfg} is eliminated.

%% file: sec/exp.tex
\section{Experiments}
\label{sec:exp}

\subsection{Experiment Setting}
\label{sec:exp_setting}

\noindent
\textbf{Pre-trained models and datasets}. Both the network parameters of student and teacher models are initialized from pre-trained diffusion models provided by EDM~\cite{karras2022edm} and LDM~\cite{rombach2022ldm}. We report quantitative as well as qualitative results on datasets with various resolutions including CIFAR10 32$\times$32~\cite{krizhevsky2009learning}, ImageNet 64$\times$64~\cite{russakovsky2015ImageNet} and latent-space LSUN-Bedroom 256$\times$256~\cite{yu2015lsun}. For Stable Diffusion~\cite{rombach2022ldm}, we use the v1.5 checkpoint and generate images with a resolution of 512$\times$512.

\noindent
\textbf{Training}. The configuration obtained in Section \ref{subsec:fast} can be applied to different NFEs and datasets. Generally, in the training of \ourName and \ourName-v, we use DPM-Solver++(3M)~\cite{lu2022dpmpp} as the teacher solver with $K=4$ (see Appendix \ref{subsec:quanti_and_quali} for an ablation study on $K$). The use of adjusted $t_{\min}=0.006$, AFS and L1 loss introduced in Section \ref{subsec:fast} all lead to improved results.
Minor changes are needed for text-to-image generation with Stable Diffusion, where we use DPM-Solver++(2M), which is the default setting used in Stable Diffusion and $K=3$. In this case, $t_{\min}$ is increased from 0.03 to 0.1 and the AFS is disabled due to the complex trajectory shown in Figure \ref{fig:salce7}. These experiment settings are collected in Table \ref{tab:exp_config} in Appendix.

\noindent
\textbf{Optimization}. We use Adam optimizer~\cite{kingma2014adam} with $\beta_1=0.9$ and $\beta_2=0.999$ and a batch size of 128 across all datasets. A learning rate of 1e-5 is used for ImageNet and LSUN-Bedroom while 5e-5 is used in other cases. We divide the learning rate by 10 halfway through training.
Our \ourName is trained with a total of 200K teacher trajectories generated (around 1.5K training iterations). We train \ourName-v to enable sampling with NFE from 2 to 5, the total training iterations is multiplied by 4 accordingly. All experiments are conducted with a maximum of 4 NVIDIA A100 GPUs. To enable a batch size of 128 using Stable Diffusion, we accumulate the gradient through several rounds.

\noindent
\textbf{Evaluation}. We measure the sample quality via Fr\'{e}chet Inception Distance (FID)~\cite{heusel2017gans} with 50K images in general. For text-to-image generation, we use a guidance scale of 7.5 to generate 5K images with prompts from the MS-COCO~\cite{lin2014microsoft} validation set. The FID is evaluated following the protocol in~\cite{liu2023instaflow,meng2023distillation,sauer2023adversarial} where the validation set serves as reference images. The CLIP score is computed using the ViT-g-14 CLIP model~\cite{radford2021learning} trained on LAION-2B~\cite{schuhmann2022laion}. 

\vspace{-1.5\baselineskip}
\begin{minipage}[t]{0.49\textwidth}
\renewcommand{\arraystretch}{0.7}
\begin{table}[H]
    \caption{Results on CIFAR10 $32\times32$.}
    \centering
    \fontsize{7}{11}\selectfont
    \begin{tabular}{lccc}
      \toprule
      \multirow{2}{*}{Method} & \multirow{2}{*}{NFE} & \multirow{2}{*}{FID} & Training time \\
      &  &  & (A100 hours) \\
      \toprule
      \multicolumn{4}{l}{\fontsize{7}{1}\selectfont \textbf{Solver-based Methods}} \\
      DDIM~\cite{song2021ddim}          & 10 & 15.69 & 0 \\
                                        & 50 & 2.91  & 0 \\
      DPM++(3M)~\cite{lu2022dpmpp}      & 5  & 24.97 & 0 \\
                                        & 10 & 3.00  & 0 \\
      AMED-Plugin~\cite{zhou2023fast}   & 5  & 6.61  & $\sim$ 0.08 \\
                                        & 10 & 2.48  & $\sim$ 0.11 \\
      GITS~\cite{chen2024trajectory}        & 5  & 8.38  & $<$ 0.01 \\
                                        & 10 & 2.49  & $\sim$ 0.01 \\
      \toprule
      \multicolumn{4}{l}{\fontsize{7}{1}\selectfont \textbf{Diffusion Distillation}} \\
      PD~\cite{salimans2022progressive}             & 1  & 9.12  & $\sim$ 195 \\
                                                    & 2  & 4.51  & $\sim$ 171 \\
      Guided PD~\cite{meng2023distillation}         & 1  & 8.34  & $\sim$ 146 \\
                                                    & 2  & 4.48  & $\sim$ 128 \\
                                                    & 4  & 3.18  & $\sim$ 119 \\
      CD~\cite{song2023consistency}                 & 1  & 3.55  & $\sim$ 1156 \\
                                                    & 2  & 2.93  & $\sim$ 1156 \\
      CTM~\cite{kim2023consistency}                 & 1  & 1.98  & $\sim$ 83 \\
      CTM~\cite{kim2023consistency} w/o GAN loss    & 1  & $>$ 5 & $\sim$ 60 \\
      \midrule
      \textbf{\ourName (ours)} (second-stage)       & 1  & 5.83  & 4.88 \\
      \textbf{\ourName (ours)}                      & 2  & 4.53  & 0.64 \\
                                                    & 3  & 3.58  & 0.92 \\
                                                    & 4  & 3.24  & 1.17 \\
                                                    & 5  & 3.06  & 1.42 \\
      \midrule
      \textbf{\ourName-v (ours)}                    & 2  & 4.28  & \multirow{4}{*}{4.26} \\
                                                    & 3  & 3.50  &  \\
                                                    & 4  & 3.18  &  \\
                                                    & 5  & 2.95  &  \\
      \bottomrule
    \end{tabular}
    \label{tab:cifar10}
\end{table}
\end{minipage}
\hfill
\begin{minipage}[t]{0.49\textwidth}
\renewcommand{\arraystretch}{0.7}
\begin{table}[H]
    \caption{Results on ImageNet $64\times64$.}
    \centering
    \fontsize{7}{11}\selectfont
    \begin{tabular}{lccc}
      \toprule
      \multirow{2}{*}{Method} & \multirow{2}{*}{NFE} & \multirow{2}{*}{FID} & Training time \\
      &  &  & (A100 hours) \\
      \toprule
      \multicolumn{4}{l}{\fontsize{7}{1}\selectfont \textbf{Solver-based Methods}} \\
      DDIM~\cite{song2021ddim}          & 10 & 16.72 & 0 \\
                                        & 50 & 4.09  & 0 \\
      DPM++(3M)~\cite{lu2022dpmpp}      & 5  & 25.49 & 0 \\
                                        & 10 & 5.67  & 0 \\
      AMED-Plugin~\cite{zhou2023fast}   & 5  & 13.83 & $\sim$ 0.18 \\
                                        & 10 & 5.01  & $\sim$ 0.32 \\
      GITS~\cite{chen2024trajectory}        & 5  & 10.79 & $<$ 0.02 \\
                                        & 10 & 4.48  & $\sim$ 0.02 \\
      \toprule
      \multicolumn{4}{l}{\fontsize{7}{1}\selectfont \textbf{Diffusion Distillation}} \\
      PD~\cite{salimans2022progressive}             & 1  & 15.39 & $<$ 5533 \\
                                                    & 2  & 8.95  & $<$ 4611 \\
      Guided PD~\cite{meng2023distillation}         & 1  & 22.74 & $<$ 5533 \\
                                                    & 2  & 9.75  & $<$ 4611 \\
                                                    & 4  & 4.14  & $<$ 4150 \\
      CD~\cite{song2023consistency}                 & 1  & 6.20  & $<$ 7867 \\
                                                    & 2  & 4.70  & $<$ 7867 \\
      CTM~\cite{kim2023consistency}                 & 1  & 2.06  & $<$ 902 \\
                                                    & 2  & 1.90  & $<$ 902 \\
      \midrule
      \textbf{\ourName (ours)} (second-stage)       & 1  & 12.89 & 6.86 \\
      \textbf{\ourName (ours)}                      & 2  & 10.25 & 3.34 \\
                                                    & 3  & 6.35  & 4.63 \\
                                                    & 4  & 4.99  & 5.98 \\
                                                    & 5  & 4.33  & 7.11 \\
      \midrule
      \textbf{\ourName-v (ours)}                    & 2  & 9.47 & \multirow{4}{*}{23.62} \\
                                                    & 3  & 5.78  &  \\
                                                    & 4  & 4.72  &  \\
                                                    & 5  & 4.21  &  \\
      \bottomrule
    \end{tabular}
    \label{tab:imagenet}
\end{table}
\end{minipage}

\subsection{Main Results}
We mainly compare our proposed \ourName and \ourName-v with progressive distillation~\cite{salimans2022progressive,meng2023distillation} and consistency distillation~\cite{song2023consistency,kim2023consistency}. 
In Table \ref{tab:cifar10} and \ref{tab:imagenet}, we report unconditional and conditional results on the pixel-space image generation. To compare the training cost, we estimate the training time measured by hours consumed on a single NVIDIA A100 GPU, following the training settings in the original papers. The detail of our estimation is included in Appendix \ref{sec:training_cost}. Our \ourName achieves comparable results as progressive distillation but only requires a very small fine-tuning cost (100$\times$ to 200$\times$ speedup). At the same time, it is hard for solver-based methods to give high-quality generation within a few steps due to increased errors. In accordance with our finding in Section \ref{subsec:smooth}, the \ourName-v shows consistently better results than \ourName, although the training cost of \ourName-v and \ourName is roughly the same for each specified sampling step. These observations also apply to the performance of \ourName and \ourName-v on the latent-space image generation on LSUN-Bedroom shown in Table \ref{tab:bedroom}. 
In Table \ref{tab:sd}, we show the performance of our methods in terms of FID and CLIP scores with a guidance scale of 7.5. The results demonstrate the effectiveness of our strategy proposed in Section \ref{subsec:guided_gtd} where the training is performed with the guidance scale set to 1. The qualitative results are shown in Figure \ref{fig:sd}. We include more results in Appendix \ref{subsec:quanti_and_quali}.

\begin{wrapfigure}{r}{0.3\textwidth}
  \vspace{-\intextsep}
  \includegraphics[width=0.3\textwidth]{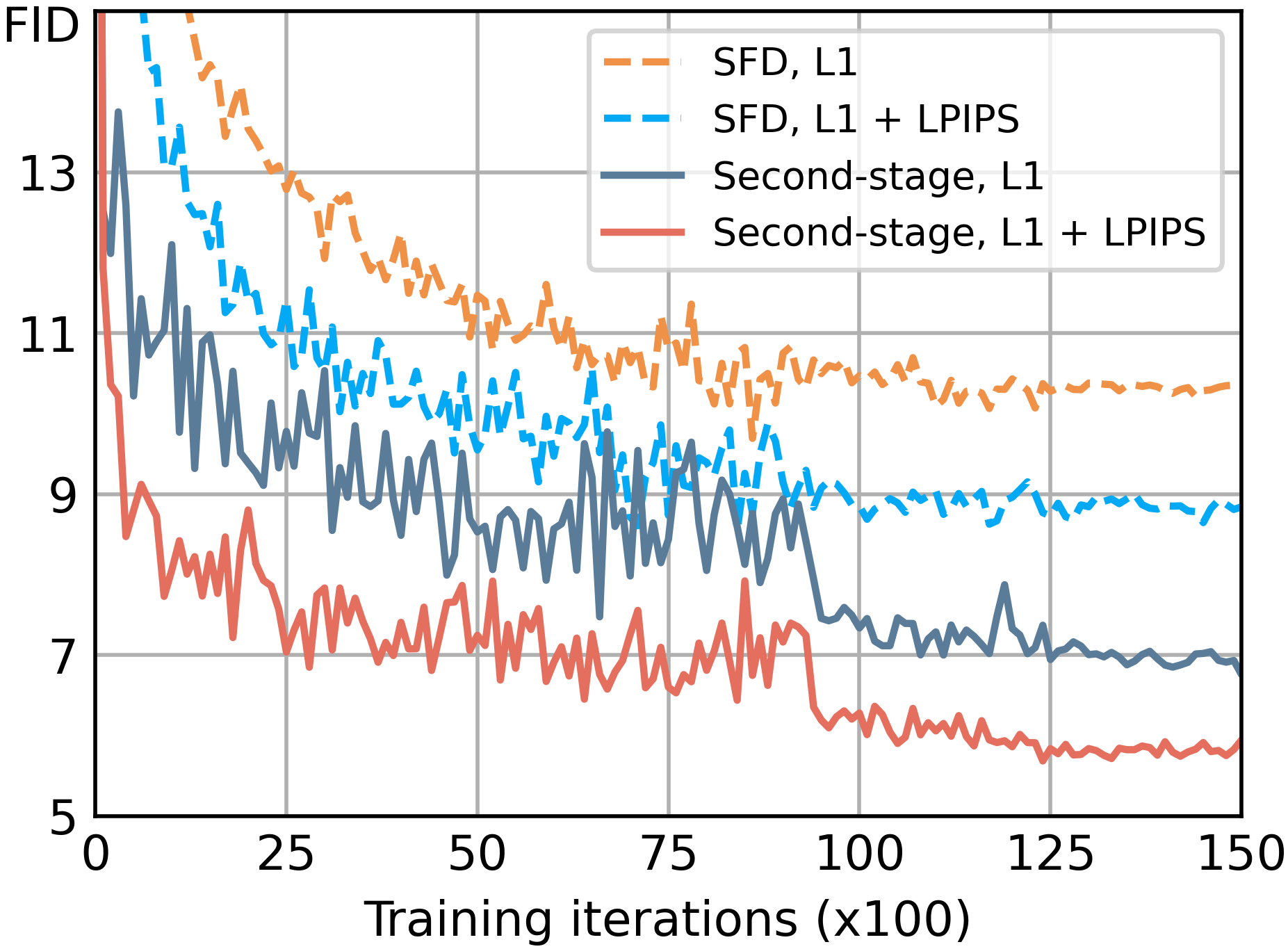}
  \caption{\it \small Ablation study on one-NFE distillation.}
  \vspace{-1\intextsep}
  \label{fig:ablation_one}
\end{wrapfigure}

Although generating images with one NFE is possible with \ourName and \ourName-v, we find it suboptimal. To address this, we propose a second-stage one-NFE distillation, initializing network parameters from a fine-tuned \ourName model. In this second stage, the teacher solver is set to DDIM (as used in \ourName), and we use AFS (\(N=2\)) and \(K=2\). The training procedure remains the same as that of \ourName. The results, marked as ``second-stage'', are reported in Tables \ref{tab:cifar10} to \ref{tab:bedroom}. We provide an ablation study on the effectiveness of the second stage and the LPIPS metric~\cite{zhang2018unreasonable} in Figure \ref{fig:ablation_one}. The second-stage training significantly boosts performance and is more efficient. Additionally, combining L1 loss with LPIPS loss yields better results. Since the teacher requires a smaller NFE, the training of each iteration of the second-stage distillation is fast. Therefore, for CIFAR10/ImageNet, we perform second-stage distillation with 2000/800K sampling trajectories (around 15/6K training iterations), and the learning rate is set to 10 times larger. For LSUN-Bedroom, we use 800K trajectories and disable the LPIPS loss.

\vspace{-1.5\baselineskip}
\begin{minipage}[t]{0.49\textwidth}
\renewcommand{\arraystretch}{0.7}
\begin{table}[H]
    \caption{Results on LSUN-Bedroom $256\times256$.}
    \centering
    \fontsize{7}{11}\selectfont
    \begin{tabular}{lcc}
      \toprule
      Method & NFE & FID \\
      \toprule
      DPM++(3M)~\cite{lu2022dpmpp}      & 8  & 4.61 \\
      AMED-Plugin~\cite{zhou2023fast}   & 8  & 4.19 \\
      PD~\cite{salimans2022progressive}             & 1  & 16.92 \\
                                                    & 2  & 8.47  \\
      CD~\cite{song2023consistency}                 & 1  & 7.80  \\
                                                    & 2  & 5.22  \\
      \textbf{\ourName (ours)} (second-stage)       & 1  & 13.88 \\
      \textbf{\ourName (ours)}                      & 2  & 10.39 \\
                                                    & 3  & 6.42  \\
                                                    & 4  & 5.26  \\
                                                    & 5  & 4.73 \\
      \textbf{\ourName-v (ours)}                    & 2  & 9.25  \\
                                                    & 3  & 5.36  \\
                                                    & 4  & 4.63  \\
                                                    & 5  & 4.33  \\
      \bottomrule
    \end{tabular}
    \label{tab:bedroom}
\end{table}
\end{minipage}
\hfill
\begin{minipage}[t]{0.49\textwidth}
\renewcommand{\arraystretch}{0.7}
\begin{table}[H]
    \caption{Text-to-image generation with Stable Diffusion v1.5~\cite{rombach2022ldm}. *: Reported in the original paper~\cite{meng2023distillation}. We use a guidance scale of 7.5, which is the default setting in the original repository.}
    \centering
    \fontsize{7}{11}\selectfont
    \begin{tabular}{lccc}
      \toprule
      Method & Steps & FID-5K & CLIP Score \\
      \midrule
      DPM++(2M)~\cite{lu2022dpmpp}      & 2  & 91.5 (98.8*) & 0.20 (0.19*) \\
                                        & 4  & 31.1 (34.1*) & 0.29 (0.29*) \\
                                        & 8  & 25.1 (25.6*) & 0.32 (0.30*) \\
      Guided PD~\cite{meng2023distillation}         & 2  & 37.3 & 0.27   \\
                                                    & 4  & 26.0 & 0.30   \\
                                                    & 8  & 26.9 & 0.30   \\
      SnapFusion~\cite{li2024snapfusion}            & 8  & 24.2 & 0.30   \\
      \textbf{\ourName-v (ours)}                    & 2  & 42.9 & 0.24  \\
                                                    & 3  & 27.6 & 0.27  \\
                                                    & 4  & 24.2 & 0.28  \\
                                                    & 5  & 23.5 & 0.29  \\
      \bottomrule
    \end{tabular}
    \label{tab:sd}
\end{table}
\end{minipage}

%% file: sec/conclusion.tex
\section{Conclusion}
\label{sec:conclusion}
In this paper, we introduce {\bf S}imple and {\bf F}ast {\bf D}istillation (\ourName) of diffusion models to achieve fast and high-quality generation with diffusion models in a few sampling steps at minimal fine-tuning cost. Through a comprehensive investigation of several important factors, we unlock \ourName's potential, achieving sample quality comparable to progressive distillation while reducing fine-tuning costs by over 100 times. To enable sampling with variable NFEs using a single distilled model, we propose \ourName-v, which incorporates step-condition as an additional input. Our methods strike a good balance between sample quality and fine-tuning costs for few-step image generation, offering a new paradigm for distillation-based accelerated sampling of diffusion models.

\noindent
\textbf{Limitation and future work}. Despite demonstrating efficient training, the FID results of our methods currently do not match those of the state-of-the-art. In future work, we plan to further explore the core mechanisms affecting the performance of trajectory distillation. Additionally, given the recent discoveries of the remarkable regular geometric structure of diffusion models~\cite{chen2024trajectory}, we aim to tailor an appropriate time schedule for our methods. We also intend to validate the effectiveness of the factors we have discussed in other distillation-based methods.

\noindent
\textbf{Broader impacts}. Similar to existing works on content creation, our methods have the potential to be misused for malicious generation, which could have harmful social impacts. However, this risk can be mitigated through advanced deepfake detection techniques. By continuously improving these detection methods, we can help ensure the responsible and ethical use of our technology.

%% file: sec/appendix.tex
\clearpage
\appendix

\input{sec/related}



\section{Estimation of Training Cost}
\label{sec:training_cost}

In this section, we illustrate our estimation of training cost on the comparison methods shown in Table \ref{tab:cifar10} and \ref{tab:imagenet}. For solver-based methods like AMED-Solver~\cite{zhou2023fast} and GITS~\cite{chen2024trajectory}, we borrow the reported training cost in original papers since similar devices are used. Efforts are put on the estimation of distillation-based methods. Due to limited resources, we do not fully re-implement these methods. Generally, after the training stabilizes, we collect the consumed time spent on several training iterations and then estimate the total training costs. Two NVIDIA A100 GPUs are used in our estimation and we double the estimated training costs to report the final results. For fair comparison, all of these methods are re-implemented with the EDM~\cite{karras2022edm} repository with its provided pre-trained diffusion models. We note that the cost of training a diffusion model on CIFAR10 dataset from scratch is around 200 A100 hours.

\subsection{Training Costs on CIFAR10}

\textbf{Progressive distillation}. Following the setting in the original paper~\cite{salimans2022progressive}, we use a batch size of 128 and strictly follows the original algorithm. Adam optimizer as well as exponential moving average (EMA) is also used. After the training stabilizes, it costs around 140.5 seconds for every 320 training iterations with 2 NVIDIA A100 GPUs. As a total of 800K training iterations are required in the original paper, we estimate its training cost by $800 \times 1000 \times 140.5 \times 2/ (320 \times 3600)=195.1$ A100 hours. As for Guided PD~\cite{meng2023distillation}, we only report the estimated training cost for its second-stage distillation, which requires a total of 600K training iterations. We roughly use the above statistics and estimate its training cost by $600 \times 1000 \times 140.5 \times 2/ (320 \times 3600)=146.4$. For the training cost under larger NFE, we scale the estimated training cost accordingly following the total number of training iterations. 

\textbf{Consistency distillation}. The consistency distillation~\cite{song2023consistency} uses a batch size of 512 and trains for a total of 800K training iterations. Following the original paper, we use the LPIPS~\cite{zhang2018unreasonable} loss metric and the Rectified Adam optimizer~\cite{liu2019variance}. It costs around 52 seconds for every 20 training iterations with 2 NVIDIA A100 GPUs. The training cost is thus estimated by $800 \times 1000 \times 52 \times 2/ (20 \times 3600)=1155.6$ A100 hours.

\textbf{Consistency trajectory models}. Following the original paper~\cite{kim2023consistency}, we use a batch size of 256 and train with mixed-precision. For the first 50K training without the GAN loss, it costs around 21.5 seconds for every 20 training iterations and 38.5 seconds are needed for the later 50K training with GAN loss involved using 2 NVIDIA A100 GPUs. The training cost of a total of 100K iterations is estimated by $50 \times 1000 \times (21.5+38.5) \times 2/ (20 \times 3600)=83.3$ A100 hours. If the GAN loss is disabled during training, the training cost will be reduced to 59.7 A100 hours.

\subsection{Training Costs on ImageNet}

Typically, the training of ImageNet uses a large batch size (for example, the original setting is 2048 for all the methods below) which requires extensive resources (usually 64 GPUs are needed). Due to limited resources, we estimate the training cost with a batch size of 256 and multiply it by 8 which should gives an upper bound of the practical training cost.

\textbf{Progressive distillation}. In progressive distillation~\cite{salimans2022progressive} as well as its guided version~\cite{meng2023distillation}, a total of 600K training iterations are required (50K for 8 rounds and 100K for 2 rounds). With a batch size of 256, it costs around 664 seconds for every 320 training iterations with 2 NVIDIA A100 GPUs. The training cost is thus $8 \times 600 \times 1000 \times 664 \times 2/ (320 \times 3600)=5533.3$ A100 hours. For the training cost under larger NFE, the training cost is scaled accordingly as done before. 

\textbf{Consistency distillation}. Following the original setting~\cite{song2023consistency}, mixed-precision optimization is applied. Other settings is similar to that in CIFAR10 illustrated above. With a batch size of 256, it costs around 59 seconds for every 20 training iterations with 2 NVIDIA A100 GPUs. As a total of 600K iterations are required, the training cost is estimated by $8 \times 600 \times 1000 \times 59 \times 2/ (20 \times 3600)=7866.7$ A100 hours.

\textbf{Consistency trajectory models}. Expect for a training iteration of 30K, other settings are consistent as in CIFAR10 training. 
For the first 10K training iterations without GAN loss, it costs around 122 seconds for every 20 training iterations with 2 NVIDIA A100 GPUs. 
For the later 20K training iterations with GAN loss, 142 seconds are needed. 
The total training cost is estimated by $8 \times 1000 \times (10\times122+20\times142) \times 2/ (20 \times 3600)=902.2$ A100 hours.

\section{Algorithms}
\label{sec:algo}

All the algorithms involved in the main text are illustrated below.

\vspace{-1.5\baselineskip}
\begin{minipage}[t]{0.49\textwidth}
\begin{algorithm}[H]
\caption{Trajectory Distillation}
  \begin{algorithmic}
  \STATE \textbf{Input:} model parameters $\psi=\theta$, learning rate $\eta$, ${\rm{Solver}}(\cdot,\cdot,\cdot,\cdot)$, distance metric $d(\cdot,\cdot)$, number of student sampling steps $N$, noise schedule $\{t_n\}_{n=0}^N$, number of teacher sampling steps between every two noise levels $K$, dataset $\mathcal{D}$.
  \REPEAT
    \STATE Sample $\bfx_0 \sim \mathcal{D}$
    \STATE Sample $n \sim \mathcal{U}(0,N-1)$
    \STATE Sample $\bfx_{n+1} \sim \mathcal{N}(\bfx_0;{t_{n+1}^2}\bfI)$
    \STATE $\bfx_n^\psi \leftarrow {\rm{Euler}}(\bfx_{n+1}, t_{n+1}, t_n, 1; \psi)$
    \STATE $\tilde{\bfx}_n \leftarrow {\rm{Solver}}(\bfx_{n+1}, t_{n+1}, t_n, K; \theta)$
    \STATE $\mathcal{L}(\psi) \leftarrow d(\bfx_n^\psi, \tilde{\bfx}_n)$
    \STATE $\psi \leftarrow \psi - \eta\nabla_\psi\mathcal{L}(\psi)$
  \UNTIL convergence
  \end{algorithmic}
\end{algorithm}
\end{minipage}
\hfill
\begin{minipage}[t]{0.49\textwidth}
\begin{algorithm}[H]
\caption{\ourName (our method)}
  \begin{algorithmic}
  \STATE \textbf{Input:} model parameters $\psi=\theta$, learning rate $\eta$, ${\rm{Solver}}(\cdot,\cdot,\cdot,\cdot)$, distance metric $d(\cdot,\cdot)$, number of student sampling steps $N$, noise schedule $\{t_n\}_{n=0}^N$, number of teacher sampling steps between every two noise levels $K$.
  \REPEAT
    \STATE Sample $\bfx_N = \tilde{\bfx}_{N} \sim \mathcal{N}(\mathbf{0};{t_{N}^2}\bfI)$
    \FOR{$n=N-1$ \textbf{to} $0$}
      \STATE $\bfx_n^\psi \leftarrow {\rm{Euler}}(\bfx_{n+1}, t_{n+1}, t_n, 1; \psi)$
      \STATE $\tilde{\bfx}_n \leftarrow {\rm{Solver}}(\tilde{\bfx}_{n+1}, t_{n+1}, t_n, K; \theta)$
      \STATE $\psi \leftarrow \psi - \eta\nabla_\psi d(\bfx_n^\psi, \tilde{\bfx}_n)$
      \STATE $\bfx_n \leftarrow {\rm{detach}}(\bfx_n^\psi)$
    \ENDFOR
  \UNTIL convergence
  \end{algorithmic}
\end{algorithm}
\end{minipage}

\begin{minipage}[t]{0.49\textwidth}
\begin{algorithm}[H]
\caption{\ourName with AFS}
  \begin{algorithmic}
  \label{alg:gtd_afs}
  \STATE \textbf{Input:} model parameters $\psi=\theta$, learning rate $\eta$, ${\rm{Solver}}(\cdot,\cdot,\cdot,\cdot)$, distance metric $d(\cdot,\cdot)$, number of student sampling steps $N$, noise schedule $\{t_n\}_{n=0}^N$, number of teacher sampling steps between every two noise levels $K$.
  \REPEAT
    \STATE Sample $\bfx_N = \tilde{\bfx}_{N} \sim \mathcal{N}(\mathbf{0};{t_{N}^2}\bfI)$
    \STATE $\hat{\bfeps} \leftarrow \bfx_N / \sqrt{1 + t_N^2}$
    \STATE $\bfx_{N-1} \leftarrow \bfx_N + (t_{N-1} - t_N) \hat{\bfeps}$
    \STATE $\tilde{\bfx}_{N-1} \leftarrow {\rm{Solver}}(\tilde{\bfx}_N, t_N, t_{N-1}, K; \theta)$
    \FOR{$n=N-2$ \textbf{to} $0$}
      \STATE $\bfx_n^\psi \leftarrow {\rm{Euler}}(\bfx_{n+1}, t_{n+1}, t_n, 1; \psi)$
      \STATE $\tilde{\bfx}_n \leftarrow {\rm{Solver}}(\tilde{\bfx}_{n+1}, t_{n+1}, t_n, K; \theta)$
      \STATE $\psi \leftarrow \psi - \eta\nabla_\psi d(\bfx_n^\psi, \tilde{\bfx}_n)$
      \STATE $\bfx_n \leftarrow {\rm{detach}}(\bfx_n^\psi)$
    \ENDFOR
  \UNTIL convergence
  \end{algorithmic}
\end{algorithm}
\end{minipage}
\hfill
\begin{minipage}[t]{0.49\textwidth}
\begin{algorithm}[H]
\caption{\ourName-v (variable-NFE)}
  \begin{algorithmic}
  \label{alg:sgtd}
  \STATE \textbf{Input:} model parameters $\psi=\theta$, learning rate $\eta$, ${\rm{Solver}}(\cdot,\cdot,\cdot,\cdot)$, distance metric $d(\cdot,\cdot)$, number of student sampling steps $N$, noise schedule $\{t_n\}_{n=0}^N$, number of teacher sampling steps between every two noise levels $K$, list $L$.
  \STATE \textbf{Initialize:} inject step-condition as a new input to the model as $\bfeps_{\psi}(\bfx, t, c, {\rm{step}}=N)$
  \REPEAT
    \STATE Sample $N \sim \mathcal{U}(L)$ and generate $\{t_n\}_{n=0}^N$
    \STATE Sample $\bfx_N = \tilde{\bfx}_{N} \sim \mathcal{N}(\mathbf{0};{t_{N}^2}\bfI)$
    \FOR{$n=N-1$ \textbf{to} $0$}
      \STATE $\bfx_n^\psi \leftarrow {\rm{Euler}}(\bfx_{n+1}, t_{n+1}, t_n, 1; \psi)$
      \STATE $\tilde{\bfx}_n \leftarrow {\rm{Solver}}(\tilde{\bfx}_{n+1}, t_{n+1}, t_n, K; \theta)$
      \STATE $\psi \leftarrow \psi - \eta\nabla_\psi d(\bfx_n^\psi, \tilde{\bfx}_n)$
      \STATE $\bfx_n \leftarrow {\rm{detach}}(\bfx_n^\psi)$
    \ENDFOR
  \UNTIL convergence
  \end{algorithmic}
\end{algorithm}
\end{minipage}

\begin{minipage}[t]{0.49\textwidth}
\begin{algorithm}[H]
\caption{Second-stage one-NFE distillation}
  \begin{algorithmic}
  \label{alg:second_stage}
  \STATE \textbf{Input:} $\psi=\theta$, $\eta$, $d(\cdot,\cdot)$, $\{t_0, t_1\}$, $K$.
  \STATE \# AFS can be used similar to Algorithm \ref{alg:gtd_afs}
  \STATE \# Step-condition can be used as Algorithm \ref{alg:sgtd}
  \REPEAT
    \STATE Sample $\bfx_1 \sim \mathcal{N}(\mathbf{0};{t_1^2}\bfI)$
    \STATE $\bfx_0^\psi \leftarrow {\rm{Euler}}(\bfx_1, t_1, t_0, 1; \psi)$
    \STATE $\bfx_0 \leftarrow {\rm{Euler}}(\bfx_1, t_1, t_0, K; \theta)$
    \STATE $\psi \leftarrow \psi - \eta\nabla_\psi d(\bfx_0^\psi, \bfx_0)$
  \UNTIL convergence
  \end{algorithmic}
\end{algorithm}
\end{minipage}
\hfill
\begin{minipage}[t]{0.49\textwidth}
\begin{algorithm}[H]
\caption{\ourName/\ourName-v sampling}
  \begin{algorithmic}
  \STATE \textbf{Input:} model parameters $\psi$, number of student sampling steps $N$, noise schedule $\{t_n\}_{n=0}^N$.
  \STATE \textbf{Initialize:} Sample $\bfx_N \sim \mathcal{N}(\mathbf{0};{t_{N}^2}\bfI)$
  \STATE \# AFS can be used similar to Algorithm \ref{alg:gtd_afs}
  \STATE \# Step-condition can be used as Algorithm \ref{alg:sgtd}
  \FOR{$n=N-1$ \textbf{to} $0$}
    \STATE $\bfx_n \leftarrow {\rm{Euler}}(\bfx_{n+1}, t_{n+1}, t_n, 1; \psi)$
  \ENDFOR
  \RETURN $\bfx_0$
  \end{algorithmic}
\end{algorithm}
\end{minipage}

\clearpage
\section{Additional Results}
\label{sec:additional}

\subsection{Inject Step-condition into the Model}
\label{subsec:step_cond}
In this section, we illustrate the detail of how to inject the step-condition into diffusion models. We take the EDM~\cite{karras2022edm} models trained on CIFAR10 dataset as an example. The modifications on models trained on other datasets are similar.

Generally, we treat step-condition the same as the timestamps input to diffusion models. The step-condition go through similar operations like timestamps to obtain the step-embedding which is then added to the time-embedding in every UNetBlock. For CIFAR10 model, we use the DDPM++ backbone proposed in~\cite{song2021sde} where the positional embedding is applied to encode the input time. The detailed modifications are shown in Algorithm \ref{alg:cifar10_network}-\ref{alg:cifar10_unetblock_step}. Note that for class-conditional models, we do not add the class-embedding to the step-embedding which we find to be suboptimal.

\vspace{-1.5\baselineskip}
\begin{minipage}[t]{0.49\textwidth}
\begin{algorithm}[H]
\caption{Original network for CIFAR10}
  \begin{algorithmic}
  \label{alg:cifar10_network}
  \STATE \hspace{-\algorithmicindent} \textbf{Class} $\rm{SongUNet}$(torch.nn.Module):
  \STATE \textbf{def} $\_\_$init$\_\_$():
  \STATE \hspace{\algorithmicindent} $\cdots$
  \STATE \hspace{\algorithmicindent} self.map$\_$noise $=$ $\rm{PositionalEmbed}$()
  \STATE \hspace{\algorithmicindent} self.map$\_$layer0 $=$ $\rm{Linear}$()
  \STATE \hspace{\algorithmicindent} self.map$\_$layer1 $=$ $\rm{Linear}$()
  \begin{tikzpicture}[remember picture, overlay]
    \node[xshift=2.2cm,yshift=-0.7cm] at (0,0)
    {\includegraphics[width=2.05\textwidth]{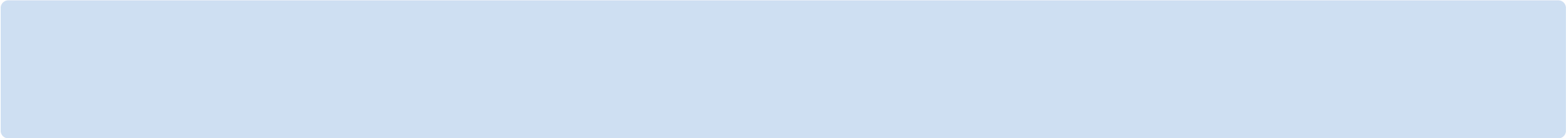}};
  \end{tikzpicture}
  \STATE
  \STATE
  \STATE
  \STATE \hspace{\algorithmicindent} $\cdots$
  \STATE
  \STATE
  \STATE \textbf{def} forward(x, noise):
  \STATE \hspace{\algorithmicindent} emb $=$ self.map$\_$noise(noise)
  \STATE \hspace{\algorithmicindent} if class$\_$conditional:
  \STATE \hspace{2\algorithmicindent} \# add class embedding to emb
  \STATE \hspace{\algorithmicindent} emb $=$ $\rm{silu}$(self.map$\_$layer0(emb))
  \STATE \hspace{\algorithmicindent} emb $=$ $\rm{silu}$(self.map$\_$layer1(emb))
  \begin{tikzpicture}[remember picture, overlay]
    \node[xshift=1.2cm,yshift=-0.7cm] at (0,0)
    {\includegraphics[width=2.05\textwidth]{pics/box_3lines.png}};
  \end{tikzpicture}
  \begin{tikzpicture}[remember picture, overlay]
    \node[xshift=1.11cm,yshift=-2.2cm] at (0,0)
    {\includegraphics[width=2.05\textwidth]{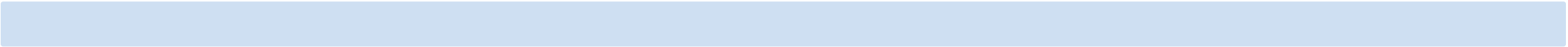}};
  \end{tikzpicture}
  \STATE
  \STATE
  \STATE
  \STATE \hspace{\algorithmicindent} $\cdots$
  \STATE \hspace{\algorithmicindent} for every $\rm{UNetBlock}$:
  \STATE \hspace{2\algorithmicindent} x $=$ $\rm{UNetBlock}$(x, emb)
  \STATE \hspace{\algorithmicindent} $\cdots$
  \end{algorithmic}
\end{algorithm}
\end{minipage}
\hfill
\begin{minipage}[t]{0.49\textwidth}
\begin{algorithm}[H]
\caption{Network with step-condition}
  \begin{algorithmic}
  \label{alg:cifar10_network_step}
  \STATE \hspace{-\algorithmicindent} \textbf{Class} $\rm{SongUNet}$(torch.nn.Module):
  \STATE \textbf{def} $\_\_$init$\_\_$():
  \STATE \hspace{\algorithmicindent} $\cdots$
  \STATE \hspace{\algorithmicindent} self.map$\_$noise $=$ $\rm{PositionalEmbed}$()
  \STATE \hspace{\algorithmicindent} self.map$\_$layer0 $=$ $\rm{Linear}$()
  \STATE \hspace{\algorithmicindent} self.map$\_$layer1 $=$ $\rm{Linear}$()
  \STATE \hspace{\algorithmicindent} self.map$\_$step $=$ $\rm{PositionalEmbed}$()
  \STATE \hspace{\algorithmicindent} self.step$\_$layer0 $=$ $\rm{Linear}$()
  \STATE \hspace{\algorithmicindent} self.step$\_$layer1 $=$ $\rm{Linear}$()
  \STATE \hspace{\algorithmicindent} $\cdots$
  \STATE
  \STATE
  \STATE \textbf{def} forward(x, noise, step):
  \STATE \hspace{\algorithmicindent} emb $=$ self.map$\_$noise(noise)
  \STATE \hspace{\algorithmicindent} if class$\_$conditional:
  \STATE \hspace{2\algorithmicindent} \# add class embedding to emb
  \STATE \hspace{\algorithmicindent} emb $=$ $\rm{silu}$(self.map$\_$layer0(emb))
  \STATE \hspace{\algorithmicindent} emb $=$ $\rm{silu}$(self.map$\_$layer1(emb))
  \STATE \hspace{\algorithmicindent} emb$\_$s $=$ self.map$\_$step(step)
  \STATE \hspace{\algorithmicindent} emb$\_$s $=$ $\rm{silu}$(self.step$\_$layer0(emb$\_$s))
  \STATE \hspace{\algorithmicindent} emb$\_$s $=$ $\rm{silu}$(self.step$\_$layer1(emb$\_$s))
  \STATE \hspace{\algorithmicindent} $\cdots$
  \STATE \hspace{\algorithmicindent} for every $\rm{UNetBlock}$:
  \STATE \hspace{2\algorithmicindent} x $=$ $\rm{UNetBlock}$(x, emb, emb$\_$s)
  \STATE \hspace{\algorithmicindent} $\cdots$
  \end{algorithmic}
\end{algorithm}
\end{minipage}

\begin{minipage}[t]{0.49\textwidth}
\begin{algorithm}[H]
\caption{Original UNetBlock for CIFAR10}
  \begin{algorithmic}
  \label{alg:cifar10_unetblock}
  \STATE \hspace{-\algorithmicindent} \textbf{Class} $\rm{UNetBlock}$(torch.nn.Module):
  \STATE \textbf{def} $\_\_$init$\_\_$():
  \STATE \hspace{\algorithmicindent} $\cdots$
  \STATE \hspace{\algorithmicindent} self.affine $=$ $\rm{Linear}$()
  \begin{tikzpicture}[remember picture, overlay]
    \node[xshift=3.1cm,yshift=-0.3cm] at (0,0)
    {\includegraphics[width=2.05\textwidth]{pics/box_1line.png}};
  \end{tikzpicture}
  \STATE
  \STATE \hspace{\algorithmicindent} $\cdots$
  \STATE
  \STATE
  \STATE \textbf{def} forward(x, emb):
  \STATE \hspace{\algorithmicindent} $\cdots$
  \STATE \hspace{\algorithmicindent} params $=$ self.affine(emb)
  \begin{tikzpicture}[remember picture, overlay]
    \node[xshift=2.45cm,yshift=-0.5cm] at (0,0)
    {\includegraphics[width=2.05\textwidth]{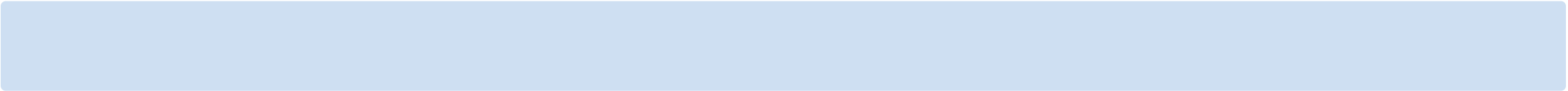}};
  \end{tikzpicture}
  \STATE
  \STATE \hspace{\algorithmicindent} x $=$ x $+$ params
  \STATE \hspace{\algorithmicindent} x $=$ $\rm{silu}$(self.norm(x))
  \STATE \hspace{\algorithmicindent} $\cdots$
  \end{algorithmic}
\end{algorithm}
\end{minipage}
\hfill
\begin{minipage}[t]{0.49\textwidth}
\begin{algorithm}[H]
\caption{UNetBlock with step-condition}
  \begin{algorithmic}
  \label{alg:cifar10_unetblock_step}
  \STATE \hspace{-\algorithmicindent} \textbf{Class} $\rm{UNetBlock}$(torch.nn.Module):
  \STATE \textbf{def} $\_\_$init$\_\_$():
  \STATE \hspace{\algorithmicindent} $\cdots$
  \STATE \hspace{\algorithmicindent} self.affine $=$ $\rm{Linear}$()
  \STATE \hspace{\algorithmicindent} self.affine$\_$s $=$ $\rm{Linear}$()
  \STATE \hspace{\algorithmicindent} $\cdots$
  \STATE
  \STATE
  \STATE \textbf{def} forward(x, emb, emb$\_$s):
  \STATE \hspace{\algorithmicindent} $\cdots$
  \STATE \hspace{\algorithmicindent} params $=$ self.affine(emb)
  \STATE \hspace{\algorithmicindent} params$\_$s $=$ self.affine(emb$\_$s)
  \STATE \hspace{\algorithmicindent} x $=$ x $+$ params $+$ params$\_$s
  \STATE \hspace{\algorithmicindent} x $=$ $\rm{silu}$(self.norm(x))
  \STATE \hspace{\algorithmicindent} $\cdots$
  \end{algorithmic}
\end{algorithm}
\end{minipage}

\begin{table}[t]
    \caption{Experiment settings used in the main text. $\dagger$: Generated teacher sampling trajectories. When training \ourName-v, it refers to the average generated trajectories used for each NFE. *: We force a batch size of 128 by accumulating the gradient for 8 rounds.}
    \label{tab:exp_config}
    \centering
    \fontsize{8}{10}\selectfont
    \begin{tabular}{lcccc}
        \toprule
        Hyperparameter & CIFAR10 & ImageNet & LSUN-Bedroom & Stable Diffusion \\
        \midrule
        Teacher solver      & DPM++(3M) & DPM++(3M) & DPM++(3M) & DPM++(2M) \\
        K                   & 4     & 4     & 4     & 3     \\
        $t_{\min}$          & 0.006 & 0.006 & 0.006 & 0.1   \\
        AFS                 & True  & True  & True  & False \\
        Generated traj.$\dagger$    & 200K  & 200K  & 200K  & 100K  \\
        Learning rate       & 5e-5  & 1e-5  & 1e-5  & 5e-5  \\
        Optimizer           & Adam  & Adam  & Adam  & Adam  \\
        Loss metric         & L1    & L1    & L1    & L1    \\
        Batch size          & 128   & 128   & 128   & 16*   \\
        Mixed-Precision     & True  & True  & True  & True  \\
        Number of GPUs      & 4     & 4     & 4     & 4     \\
        \bottomrule
    \end{tabular}
\end{table}

\begin{table}[t]
    \caption{FID results on CIFAR-10. The lines with gray background are the results reported in the main text. $\dagger$: Second-stage one-NFE distillation. For each row of the results of \ourName-v, the number of reported results corresponds to the the length of the list of sampling steps $L$ in Algorithm \ref{alg:sgtd}.}
    \label{tab:add_cifar10}
    \centering
    \fontsize{8}{10}\selectfont
    \begin{tabular}{lcccccccc}
        \toprule
        \multirow{2}{*}{Method} & \multicolumn{8}{c}{NFE} \\
        \cmidrule{2-9}
        & 1 & 2 & 3 & 4 & 5 & 6 & 7 & 8 \\
        \midrule
        \ourName (100K)     & 21.14 & 4.57 & 3.66 & 3.26 & 3.06 & 2.97 & 2.87 & 2.85 \\
        \rowcolor[gray]{0.9} \ourName (200K)      & 5.83$\dagger$ & 4.53 & 3.58 & 3.24 & 3.06 & - & - & - \\
        \rowcolor[gray]{0.9} \ourName-v (800K)    & -    & 4.28 & 3.50 & 3.18 & 2.95 & - & - & - \\
        \ourName-v (800K)   & 18.69 & 4.34 & 3.58 & 3.22 & 2.97 & 2.94 & 2.88 & 2.87 \\
        \ourName-v (2000K)  & 11.35 & 4.16 & 3.44 & 3.11 & 2.95 & - & - & - \\
        \ourName-v (2000K, conditional)    & 9.17  & 3.45 & 2.85 & 2.76 & 2.63 & - & - & - \\  
        \bottomrule
    \end{tabular}
\end{table}

\begin{table}[H]
    \caption{Ablation study on the number of teacher sampling steps $K$ on CIFAR10 dataset. We report pairs of FID and fine-tuning time (A100 hours).}
    \label{tab:ablation_K}
    \centering
    \fontsize{8}{10}\selectfont
    \begin{tabular}{lcccccc}
        \toprule
        \multirow{2}{*}{NFE} & \multicolumn{6}{c}{$K$} \\
        \cmidrule{2-7}
        & 1 & 2 & 3 & 4 & 5 & 6 \\
        \midrule
        2   & 64.11/0.17 & 18.61/0.22 & 6.51/0.28 & 4.57/0.32 & 4.99/0.38 & 5.66/0.42 \\
        3   & 40.15/0.26 & 7.76/0.32 & 3.60/0.39 & 3.66/0.46 & 3.84/0.53 & 3.97/0.59 \\
        4   & 25.96/0.33 & 5.21/0.42 & 3.34/0.49 & 3.26/0.59 & 3.32/0.67 & 3.30/0.75 \\
        5   & 16.39/0.41 & 3.78/0.51 & 3.10/0.61 & 3.06/0.71 & 3.09/0.82 & 3.17/0.92 \\
        \bottomrule
    \end{tabular}
\end{table}

\subsection{Additional Quantitative and Qualitative results}
\label{subsec:quanti_and_quali}
In Table \ref{tab:exp_config}, we include all the hyperparameters used in our main experiments (Section \ref{sec:exp}). In Table \ref{tab:add_cifar10}, we report more quantitative results on CIFAR10 dataset~\cite{krizhevsky2009learning}. By using ``100K'', we mean a total of 100K sampling trajectories are generated by the teacher model, which equals around 781 training iterations with a batch size of 128. The experiment settings here are basically in accordance with Table \ref{tab:exp_config}. For each row of the results of \ourName-v, the number of reported results corresponds to the the length of the list of sampling steps $L$ in Algorithm \ref{alg:sgtd}. For example, in the third row, the \ourName-v is trained to sample with 2, 3, 4 and 5 NFE (each NFE is trained for 200K sampling trajectories on average), while in the fourth row it is trained to sample with 1 to 8 NFE (100 K for each NFE on average). In the last row, we also include results of conditional image generation on CIFAR10 with pre-trained model provided by EDM~\cite{karras2022edm}. 

Following the final setting in Section \ref{subsec:fast}, we provide an ablation study on the intermediate teacher sampling steps $K$ in Table \ref{tab:ablation_K}. It is shown that $K=4$ achieves a good trade-off between FID and fine-tuning time.

During our experiments, we adhere to the time schedules utilized in previous studies (for instance, a polynomial schedule with $\rho=7$ for pixel-space pre-trained models from EDM~\cite{karras2022edm} and a linear schedule for latend-space models from LDM~\cite{rombach2022ldm}), and find them effective. In Figure \ref{tab:ablation_schedule}, we show an ablation study on CIFAR10 dataset with 2-NFE \ourName trained with different polynomial coefficients.

\begin{table}[H]
  \caption{Ablation study on time schedule on CIFAR10 dataset.}
  \label{tab:ablation_schedule}
  \centering
  \fontsize{8}{10}\selectfont
  \begin{tabular}{llc}
    \toprule
    $\rho$ & Time schedule & FID \\
    \midrule
    5  & [80.00, 15.11, 1.22, 0.006] & 5.50 \\
    6  & [80.00, 12.63, 0.86, 0.006] & 4.61 \\
    7  & [80.00, 10.93, 0.67, 0.006] & 4.53 \\
    8  & [80.00,  9.72, 0.55, 0.006] & 4.54 \\
    9  & [80.00,  8.82, 0.47, 0.006] & 4.60 \\
    10 & [80.00,  8.13, 0.42, 0.006] & 4.81 \\
    \bottomrule
  \end{tabular}
\end{table}

For further evaluation on fidelity and diversity, we compute precision, recall, density and coverage following standard practice~\cite{naeem2020reliable} on CIFAR10 dataset. We use the same random seed for a fair comparison. The results are shown in Table \ref{tab:eval}. In general, while achieving considerable acceleration on image generation, our method does not sacrifice diversity.

\begin{table}[H]
  \caption{Evaluation on fidelity and diversity on CIFAR10 dataset.}
  \label{tab:eval}
  \centering
  \fontsize{8}{10}\selectfont
  \begin{tabular}{lcccccc}
    \toprule
    Method & NFE & FID & Precision & Recall & Density & Coverage \\
    \midrule
    \textbf{SFD-v (ours)}        &  2 & 4.28 & 0.77 & 0.70 & 1.06 & 0.93 \\
                                 &  3 & 3.50 & 0.78 & 0.71 & 1.10 & 0.94 \\
                                 &  4 & 3.18 & 0.79 & 0.71 & 1.13 & 0.94 \\
                                 &  5 & 2.95 & 0.79 & 0.71 & 1.15 & 0.95 \\
    DPM++(3M)~\cite{lu2022dpmpp} & 11 & 3.93 & 0.76 & 0.71 & 1.04 & 0.94 \\
                                 & 15 & 2.64 & 0.76 & 0.73 & 1.03 & 0.95 \\
                                 & 19 & 2.54 & 0.77 & 0.72 & 1.04 & 0.96 \\
                                 & 23 & 2.65 & 0.77 & 0.72 & 1.05 & 0.96 \\
                                 & 50 & 2.01 & 0.78 & 0.72 & 1.11 & 0.96 \\
    DDIM~\cite{song2021ddim}     & 50 & 2.91 & 0.79 & 0.71 & 1.09 & 0.95 \\
    Heun~\cite{karras2022edm}    & 50 & 1.96 & 0.79 & 0.72 & 1.10 & 0.96 \\
    \bottomrule
  \end{tabular}
\end{table}

\begin{figure}[t]
  \includegraphics[width=\textwidth]{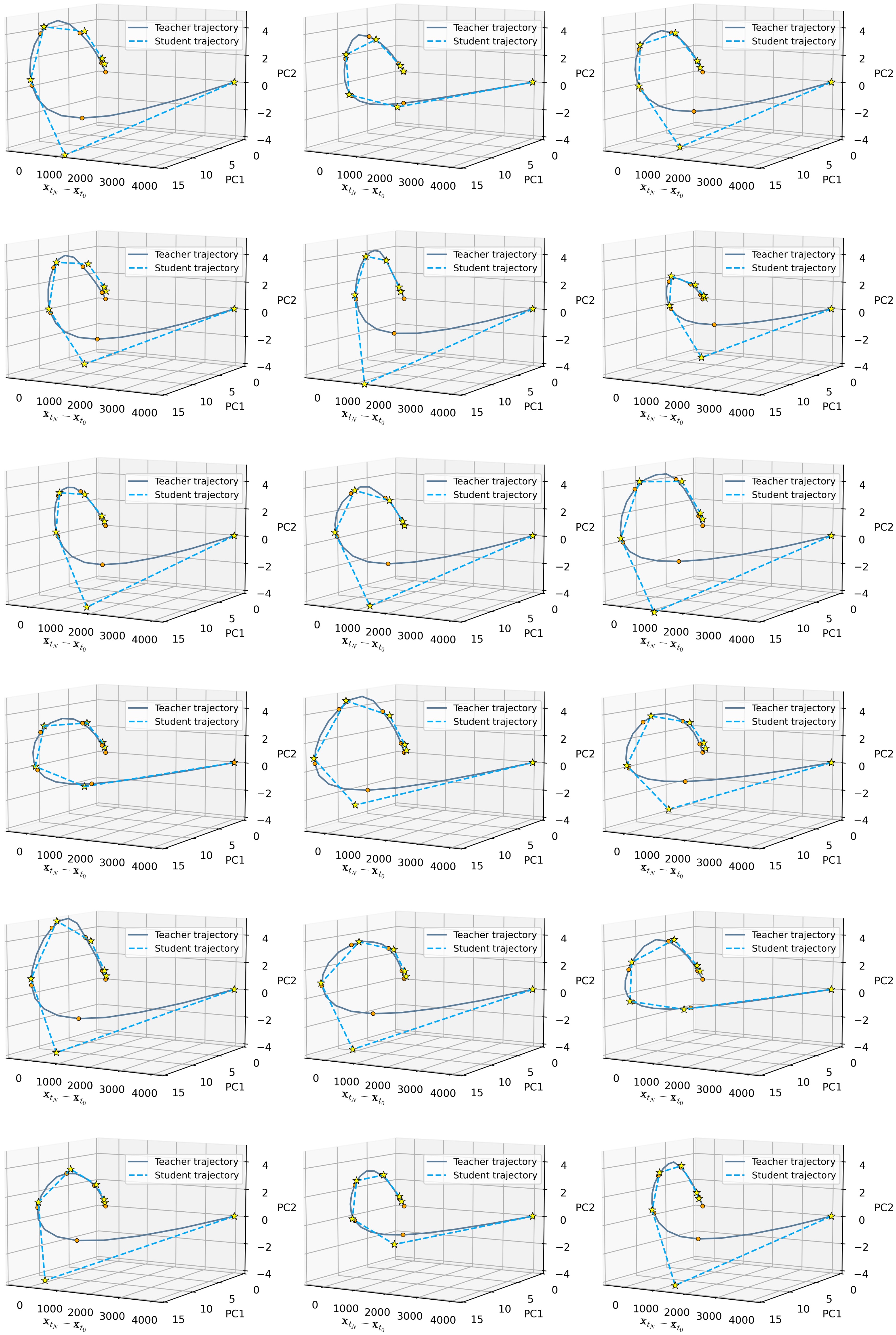}
  \caption{Visualization of the effectiveness of \ourName.}
  \label{fig:app_visial}
\end{figure}

\begin{figure}[t]
  \includegraphics[width=\textwidth]{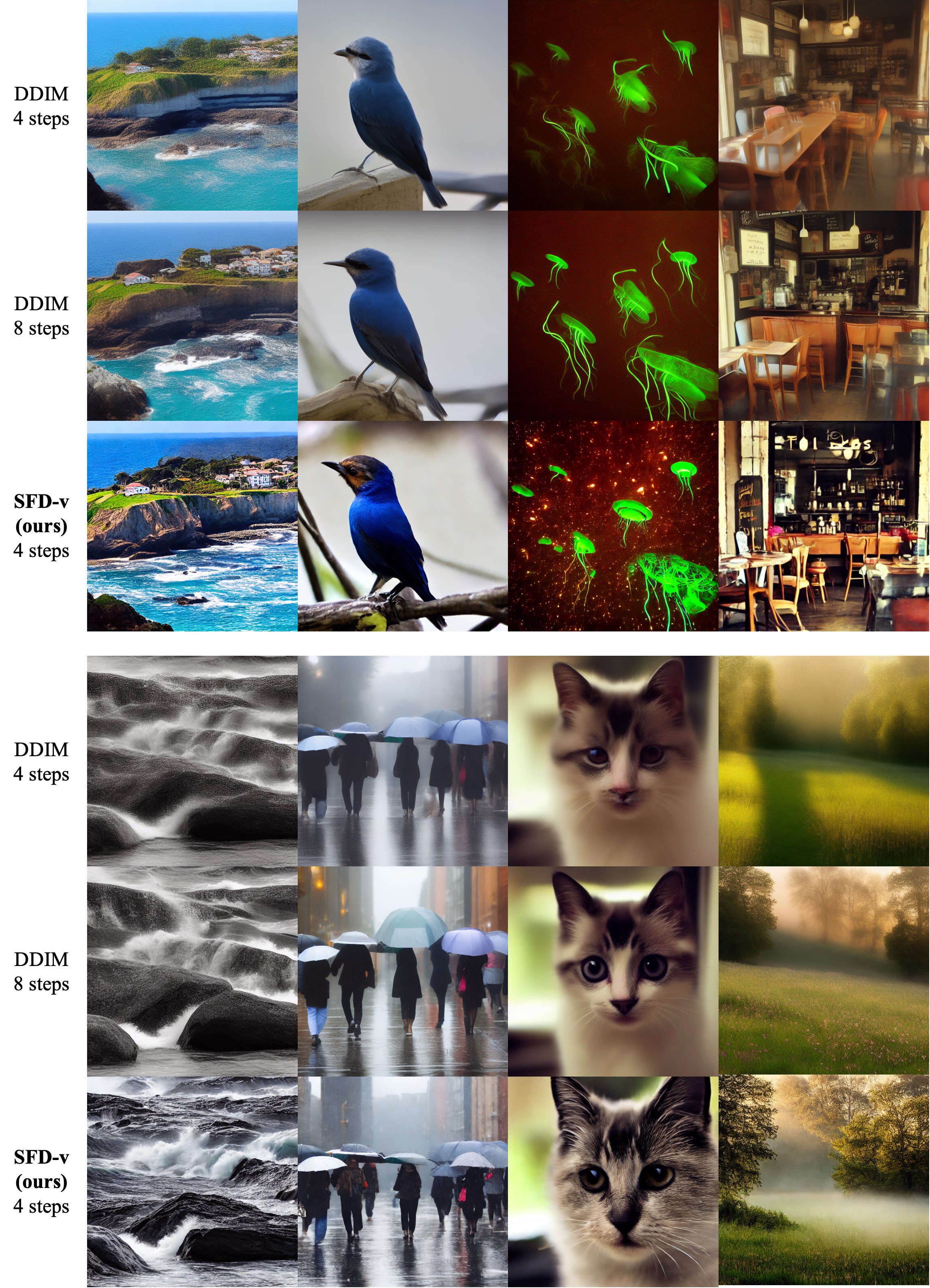}
  \caption{Qualitative results generated by Stable Diffusion v1.5~\cite{rombach2022ldm}.}
  \label{fig:qualitative_sd1}
\end{figure}

\begin{figure}[t]
  \includegraphics[width=\textwidth]{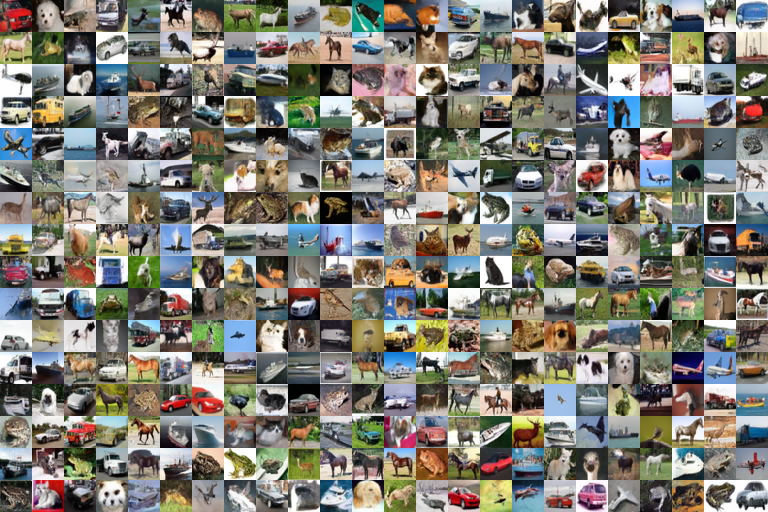}
  \caption{Uncurated qualitative results on CIFAR10. NFE=1.}
  \label{fig:qualitative_cifar10_nfe1}
\end{figure}

\begin{figure}[t]
  \includegraphics[width=\textwidth]{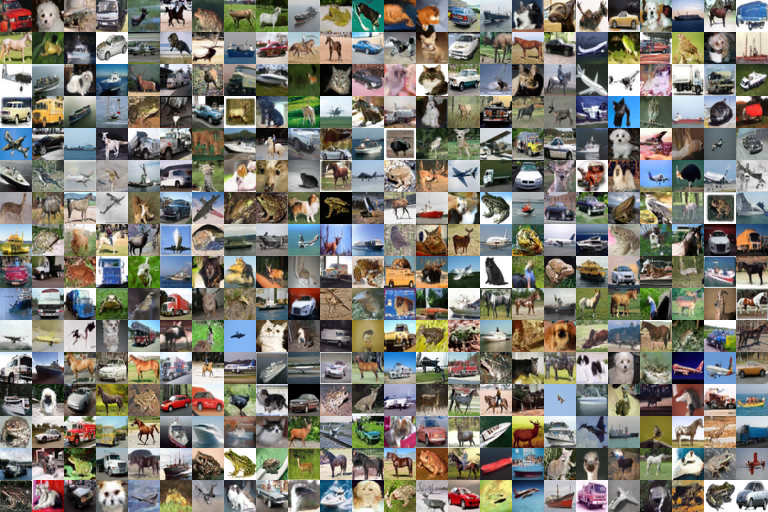}
  \caption{Uncurated qualitative results on CIFAR10. NFE=3.}
  \label{fig:qualitative_cifar10_nfe3}
\end{figure}

\begin{figure}[t]
  \includegraphics[width=\textwidth]{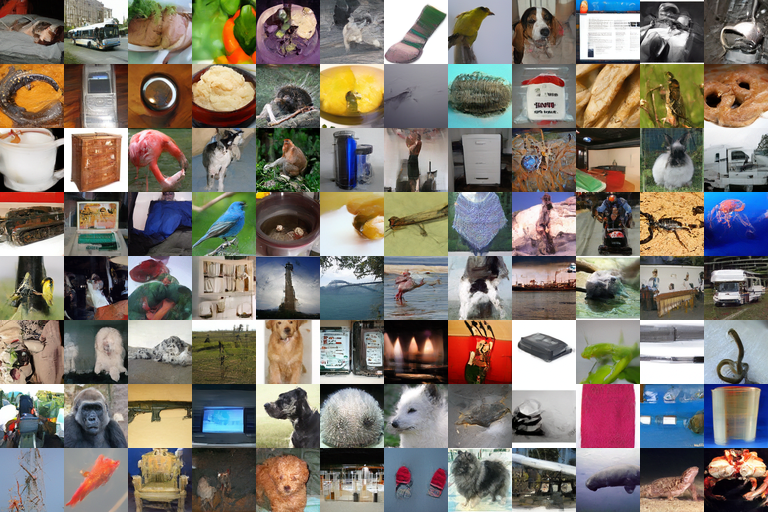}
  \caption{Uncurated qualitative results on ImageNet. NFE=1.}
  \label{fig:qualitative_imagenet_nfe1}
\end{figure}

\begin{figure}[t]
  \includegraphics[width=\textwidth]{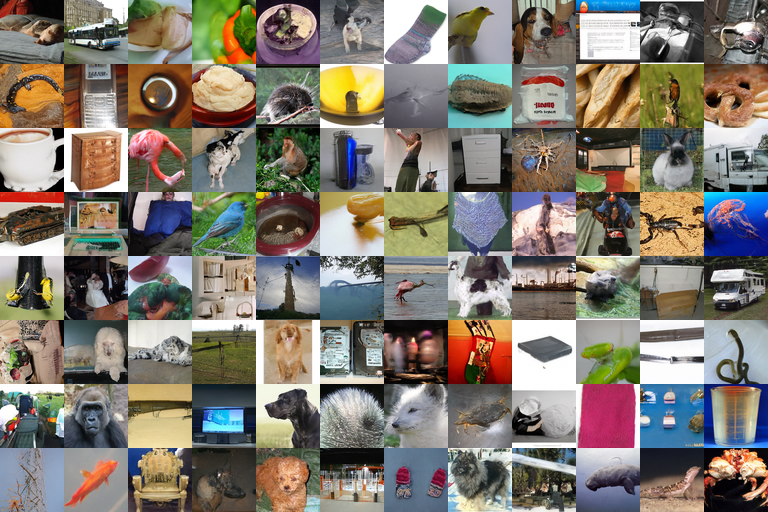}
  \caption{Uncurated qualitative results on ImageNet. NFE=3.}
  \label{fig:qualitative_imagenet_nfe3}
\end{figure}

\begin{figure}[t]
  \includegraphics[width=\textwidth]{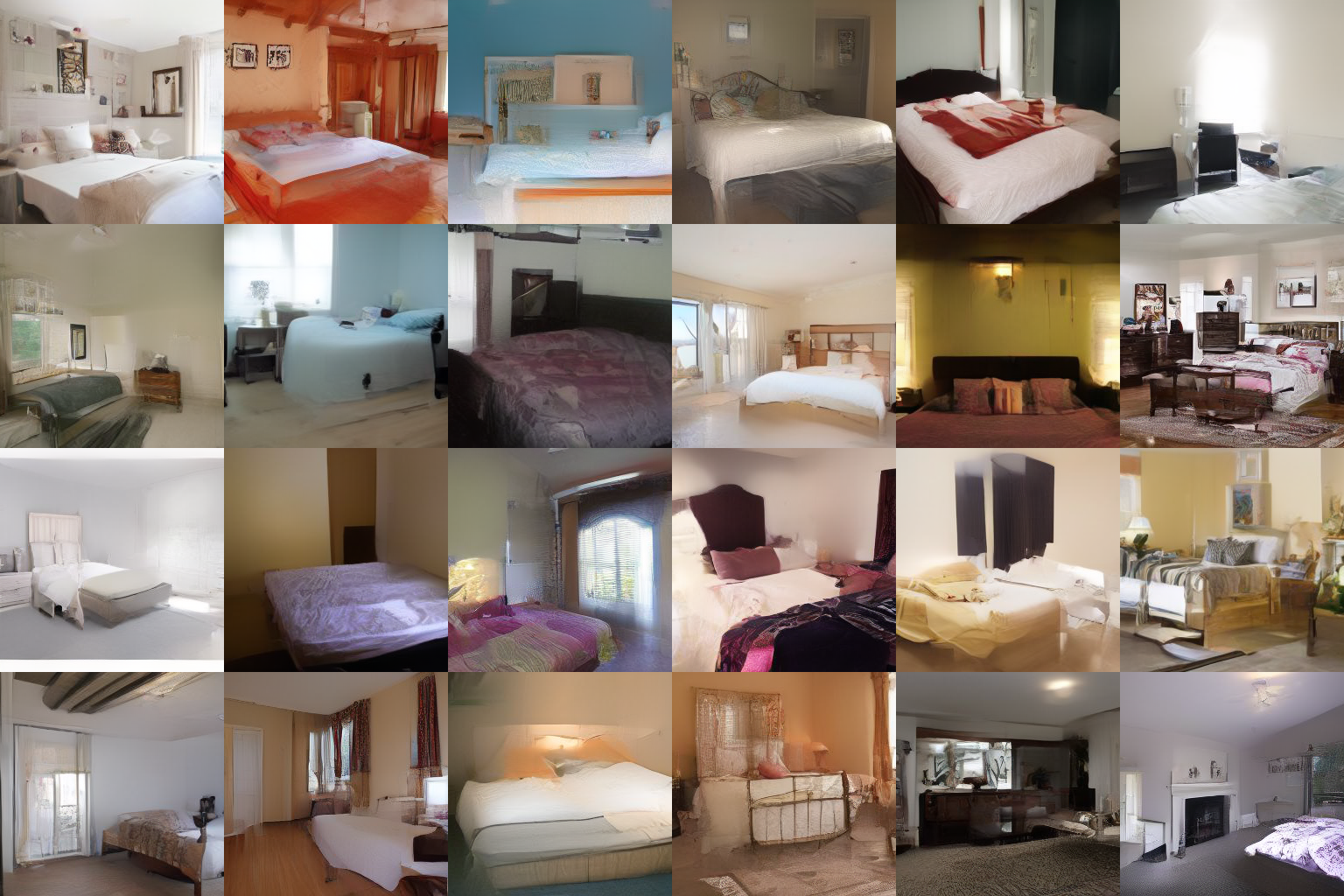}
  \caption{Uncurated qualitative results on LSUN-Bedroom. NFE=1.}
  \label{fig:qualitative_bedroom_nfe1}
\end{figure}

\begin{figure}[t]
  \includegraphics[width=\textwidth]{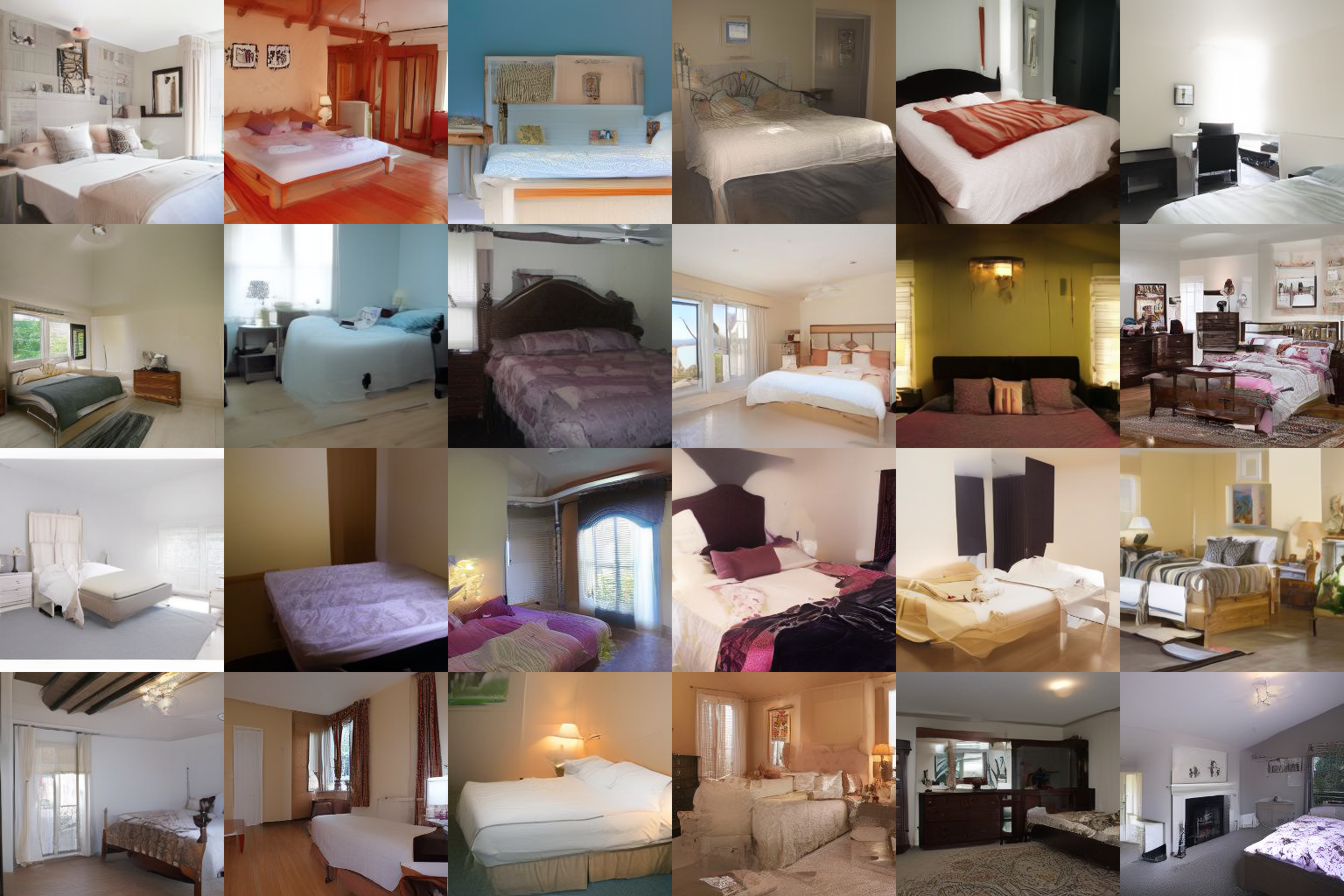}
  \caption{Uncurated qualitative results on LSUN-Bedroom. NFE=3.}
  \label{fig:qualitative_bedroom_nfe3}
\end{figure}

We include more qualitative results from Figure \ref{fig:app_visial} to Figure \ref{fig:qualitative_bedroom_nfe3}.

\clearpage

\clearpage

\clearpage

\clearpage

%% file: sec/related.tex
\section{Related Works}
\label{sec:related}

\noindent
\textbf{Solver-based methods}. As sampling from diffusion models can be interpreted as solving the PF-ODE~\cite{song2021sde}, various kinds of training-free solver-based methods are designed utilizing classical numerical methods including Euler's method (DDIM~\cite{song2021ddim}), Heun's second method (EDM~\cite{karras2022edm}), linear multi-step method (PNDM~\cite{liu2022pseudo} and iPNDM~\cite{zhang2023deis}) and predictor-corrector framework (UniPC~\cite{zhao2023unipc}). 
Some works focus on the semi-linear structure of the PF-ODE, approximating its solution by Taylor expansion (DPM-Solver~\cite{lu2022dpm}, DPM-Solver++~\cite{lu2022dpmpp} and their generalized variant SEEDS~\cite{gonzalez2024seeds}) and polynomial extrapolation (DEIS~\cite{zhang2023deis}).
Besides these training-free methods, some works further reduce the discretization error of classical numerical methods by learning intrinsic information with extra computational overhead. 
GENIE~\cite{dockhorn2022genie} applies the second-order truncated Taylor method and utilizes the gradient of noise-prediction model w.r.t. time distilled from the pre-trained model for accelerated sampling.
DPM-Solver-v3~\cite{zheng2023dpm} seeks to search for the optimal parameterization for fast sampling using the Empirical Model Statistics (EMS) calculated from pre-trained models.
AMED-Solver~\cite{zhou2023fast} achieves fast sampling resorting to the mean value theorem validated by the geometric structure of sampling trajectories and trains a small network that predicts the optimal intermediate timestamps. 
GITS~\cite{chen2024trajectory} optimizes the time schedule through a dynamic programming utilizing the trajectory regularity of the diffusion models. 

Aside from solver-based methods, distillation-based methods demonstrate their superiority in sampling speed and high-quality generation. Current literature can be categorized into three classes.

\noindent
\textbf{Trajectory distillation}. Trajectory distillation originates from the primary work~\cite{luhman2021knowledge}, which proposed the first one-step diffusion model with the idea of knowledge distillation. The basic framework behind is to train a student model to imitate the teacher's sampling trajectory. 
Progressive distillation (PD)~\cite{salimans2022progressive,meng2023distillation,li2024snapfusion} proposes gradually reducing the sampling steps with a multi-stage strategy. In each training round, the student model is fine-tuned to merge two DDIM steps into one step and serves as the teacher in the next training round. 
Following PD, trajectory matching at feature space utilizing a pre-trained classifier is shown to be effective in RCFD~\cite{sun2023accelerating} and RDD~\cite{feng2024relational}.
Motivated by the idea of operator learning, DSNO~\cite{zheng2023fast} presents a novel way of distillation by simultaneously predicting the whole sampling trajectory with specially designed temporal convolution blocks.

\noindent
\textbf{Consistency distillation}. Originated from consistency models~\cite{song2023consistency}, which is a special case of its concurrent consistent diffusion models~\cite{daras2023consistent}, consistency distillation introduces a new way of distillation where the denoising outputs on the sampling trajectory are kept consistent. Consistency distillation shows remarkable ability in one-step generation and has become popular in distillation-based methods. Latent consistency models~\cite{luo2023latent} demonstrate the effectiveness of consistency distillation in latent space. Consistency trajectory models~\cite{kim2023consistency} generalize the consistency to any timestamp, enabling an unrestricted traversal on sampling trajectory and producing state-of-the-art results. 

\noindent
\textbf{Distribution matching}. The idea of distribution matching for distilling diffusion models is first introduced in text-to-3D generation~\cite{poole2022dreamfusion} as score distillation sampling (SDS) and is later improved to variational score distillation (VSD)~\cite{wang2024prolificdreamer}. 
Unlike the training of diffusion models, which relies on sample-wise reconstruction, distribution matching discards the coupling of noise-image pairs set by diffusion models and matches the real and reconstructed samples at a distribution level. The effectiveness of both SDS~\cite{sauer2023adversarial} and VSD~\cite{yin2023one,nguyen2023swiftbrush} is recently validated in image generation.

Our methods fall into the first category but differ from progressive distillation in three ways. 
(i) We view trajectory distillation from a global perspective, where we generate the whole teacher sampling trajectory in each iteration, and let the student model imitate it step by step. 
(ii) We simplify the multi-stage strategy (more than ten stages) into one or two stages. 
(iii) The proposed \ourName-v enables sampling with different NFEs using a single distilled model.
Our methods differ from DSNO in two folds. 
(i) Our methods enable the student model to reduce accumulated errors in previous steps, while in DSNO, the intermediate samples on the predicted sampling trajectory are not directly connected. 
(ii) Our methods slightly change the network architecture, while in DSNO, it is largely redesigned. Besides, the model complexity in DSNO increases as the length of the sampling trajectory increases. 